\documentclass{article}
\pdfoutput=1

\PassOptionsToPackage{round}{natbib}

\usepackage[final]{neurips_2021}

\usepackage{hyperref} %[colorlinks=true]{hyperref}
\usepackage{color}
\hypersetup{
  colorlinks,
  linkcolor={magenta}, %!50!black},
  citecolor={blue}, %!50!black},
  urlcolor={blue}, %!80!black}
}

\usepackage{url}

\usepackage{wrapfig}

\usepackage{subcaption}
\usepackage{amsmath}

\usepackage{natbib}

\usepackage{algorithm}
\usepackage{algorithmic}
\usepackage{comment}
\usepackage{enumitem}

\usepackage{amsmath,amssymb}
\usepackage{amsbsy}
\usepackage{bm}
\usepackage{amsthm}
\usepackage{mathtools}

\newcommand{\argmin}{\mathop{\mathrm{argmin}}}

\newcommand{\Eqref}[1]{Eq. \eqref{#1}}

\newcommand{\boldone}{{\boldsymbol{1}}}

\newcommand{\calA}{{\mathcal A}}
\newcommand{\calB}{{\mathcal B}}

\newcommand{\calE}{{\mathcal E}}
\newcommand{\calF}{{\mathcal F}}

\newcommand{\calM}{{\mathcal M}}
\newcommand{\calN}{{\mathcal N}}

\newcommand{\calH}{{\mathcal H}}
\newcommand{\calX}{{\mathcal X}}

\newcommand{\fhat}{\widehat{f}}
\newcommand{\fcheck}{\check{f}}

\newcommand{\fbar}{\bar{f}}

\newcommand{\Px}{P_{\calX}}

\newcommand{\LPi}{L^2} %(P_{\calX})}
\newcommand{\LPiPx}{L^2(P_X)}

\newcommand{\Real}{\mathbb{R}}
\newcommand{\Natural}{\mathbb{N}}
\newcommand{\Integer}{\mathbb{Z}}

\newcommand{\EE}{\mathrm{E}}
\newcommand{\dd}{\mathrm{d}}
\newcommand{\fstar}{f^{\ast}}

\newcommand{\ftrue}{f^{\mathrm{o}}}

\def\I<#1>{\left\langle #1 \right\rangle}
\def\i<#1>{\left\langle #1 \right\rangle}

\newcommand{\Well}[1]{\mathcal{W}^{(#1)}}
\newcommand{\bell}[1]{b^{(#1)}}

\newcommand{\betamin}{\underline{\beta}}%{\beta_{\min}}
\newcommand{\betamax}{\overline{\beta}}%{\beta_{\max}}
\newcommand{\betatild}{\widetilde{\beta}}
\newcommand{\normb}{\|_{\betamin/\beta}}

\newcommand{\calHaff}{\calH_{\mathrm{aff}}}
\newcommand{\calHdeep}{\calH_{\mathrm{deep}}}

\newcommand{\dtil}{\tilde{d}}
\newcommand{\betast}{\beta^*}

\newcommand{\convbar}{\overline{\mathrm{conv}}}

\allowdisplaybreaks
\newtheorem{Theorem}{Theorem}
\newtheorem{Lemma}{Lemma}
\newtheorem{Definition}{Definition}

\newtheorem{Proposition}{Proposition}

\newtheorem{Corollary}{Corollary}

\newcommand{\mell}{m_\ell}
\newcommand{\melle}[1]{m_{#1}}

\newcommand{\Id}{\mathrm{I}}

\newcommand{\Holder}{H\"{o}lder }

\newcommand{\Otilde}{\tilde{O}}

\newcommand{\calC}{\mathcal{C}}

\newcommand{\MB}{M\!B}

\usepackage{enumitem}
\newlist{assumenum}{enumerate}{1} % also creates a counter called 'assumenum'
\setlist[assumenum]{label={\rm \bf (A\arabic*)}, ref=(A\arabic*),resume}%ref=\thetheorem~(\roman*)}
\setlist[assumenum,1]{leftmargin=30pt,resume}%

\title{Deep learning is adaptive to intrinsic dimensionality of model smoothness in anisotropic Besov space}

\author{%
Taiji Suzuki \\
Department of Mathematical Informatics, The University of Tokyo, Tokyo, Japan \\
RIKEN Center for Advanced Intelligence Project, Tokyo, Japan \\
  \texttt{taiji@mist.i.u-tokyo.ac.jp} \\
\And 
Atsushi Nitanda \\
Kyushu Institute of Technology, Fukuoka, Japan \\
RIKEN Center for Advanced Intelligence Project, Tokyo, Japan \\
  \texttt{nitanda@ai.kyutech.ac.jp} 
}

\begin{document}

\maketitle

\begin{abstract}
Deep learning has exhibited superior performance for various tasks, especially for high-dimensional datasets, such as images. 
To understand this property, we investigate the approximation and estimation ability of deep learning on {\it anisotropic Besov spaces}.
The anisotropic Besov space is characterized by direction-dependent smoothness and includes several function classes that have been investigated thus far.
We demonstrate that the approximation error and estimation error of deep learning only depend on the average value of the smoothness parameters in all directions. Consequently, the curse of dimensionality can be avoided if the smoothness of the target function is highly anisotropic.
Unlike existing studies, our analysis does not require a low-dimensional structure of the input data.
We also investigate the minimax optimality of deep learning and compare its performance with that of the kernel method (more generally, linear estimators).
The results show that deep learning has better dependence on the input dimensionality if the target function possesses anisotropic smoothness, and it achieves an adaptive rate for functions with spatially inhomogeneous smoothness.

%Deep learning has shown superior performances on various tasks, especially, on high dimensional datasets such as images. 
%To understand this property, we investigate approximation and estimation ability of deep learning on {\it anisotropic Besov spaces}.
%The anisotropic Besov space is characterized by a ``direction dependent'' smoothness and includes several function classes investigated so far.
%We show that the approximation error and estimation error of deep learning only depend on the average of the smoothness parameters of all directions, and consequently, it can avoid the curse of dimensionality if the smoothness of the target function is highly anisotropic.
%Unlike existing researches, our analysis does not require a low dimensional structure of the input data.
%We also investigate the minimax optimality of deep learning and compare its performance with that of the kernel method (more generally, linear estimators).
%As a result, it is shown that deep learning has better dependency on the input dimensionality if the target function has anisotropic smoothness, and achieves adaptive rate for functions with spatially inhomogeneous smoothness.
%%Our analysis is considerably general and includes several existing results.
\end{abstract}

\section{Introduction}

Based on the recent literature pertaining to machine learning, deep learning has exhibited superior performance in several tasks
such as image recognition \citep{krizhevsky2012imagenet}, natural language processing \citep{2018arXiv181004805D},
and image synthesis \citep{2015arXiv151106434R}. 
In particular, its superiority is remarkable for complicated and high-dimensional data like images. 
This is mainly due to its high flexibility and superior feature-extraction ability for effectively extracting the intrinsic structure of data.
%Apart from its successful applications
Its theoretical analysis also has been extensively developed 
considering several aspects such as expressive ability, optimization, and generalization error.
%In this paper, we theoretically analyze the expressive power and statistical property of deep learning.

%It is well known that deep neural networks have universal approximation ability %\citep{cybenko1989approximation,hornik1991approximation,sonoda2015neural}, that is,
%a deep neural network is capable of approximating any function, such as continuous functions, with any precision (e.g., with respect to the uniform %norm) by increasing the width. 
%This indicates high flexibility of the deep neural network models.
%In terms of efficiency in approximating a target function,
%the representation ability of deep neural networks increases exponentially with the number of layers 
%\citep{NIPS2014_5422,bianchini2014complexity,cohen2016expressive,ICML:Cohen+Shashua:2016,NIPS:Poole+etal:2016}.

Amongst representation ability analysis of deep neural networks such as universal approximation ability \citep{cybenko1989approximation,hornik1991approximation,sonoda2015neural},
approximation theory of deep neural networks on typical function classes such as H\"{o}lder, Sobolev, and Besov spaces have been extensively studied.
In particular, analyses of deep neural networks with the ReLU activation \citep{nair2010rectified,glorot2011deep} have been recently developed.
\citet{2017arXiv170806633S} showed that the deep learning with ReLU activations can achieve the minimax optimal estimation accuracy to estimate  composite functions in \Holder spaces by using the approximation theory of \citet{NN:Yarotsky:2017}.
% analyzed the \Holder space and derived the approximation error of deep neural networks. 
%This approximation error analysis was applied to derive the estimation error of deep leaning in estimating composite functions in \Holder spaces by \cite{2017arXiv170806633S}. They showed that ReLU deep neural networks can achieve the minimax optimal rate for that function class.
\citet{suzuki2018adaptivity} generalized this analysis to those on the {\it Besov space} and the {\it mixed smooth Besov space} by utilizing the techniques developed in approximation theories \citep{Book:Temlyakov:1993,DeVore1998nonlinearapprox}.
It was shown that deep learning can achieve an {\it adaptive approximation} error rate that is faster than that of (non-adaptive) linear approximation methods \citep{devore1988interpolation,devore1993wavelet,dung2011optimal}, and 
it outperforms any linear estimators (including kernel ridge regression) in terms of the minimax optimal rate.
%One of the key notion for the approximation theory in the Besov space and related spaces such as \Holder and Sobolev spaces is its connection to wavelet analysis \citep{Mallat99a}. \cite{bolcskei2017optimal} presented a general strategy for analyzing approximation ability of neural networks connecting with the wavelet analysis.
%Other related work includes an approximation error analysis on the class of piece-wise smooth functions \citep{petersen2017optimal} and its application to estimation error analysis on that function class \citep{arXiv:Imaizumi+Fukumizu:2018}. 
%Additional classical results for smoother activation functions have been developed previously, for example, by \citet{mhaskar1992approximation,mhaskar1993approximation,chui1994neural,mhaskar1996neural,pinkus1999approximation}.
%also developed an approximation error analysis which essentially leads to analyses for Besov spaces.

\begin{table*}[t]
\centering
\caption{Relationship between existing research and our work. $\beta$ indicates the smoothness of the target function, $d$ is the dimensionality of input $x$, $D$ is the dimensionality of a low-dimensional structure on which the data are distributed, and $\betatild$ is the average smoothness of an anisotropic Besov space (\Eqref{eq:betatilddef}).}
\label{tab:RelationBetWork}
\vspace{-0.2cm}
{\small
\begin{tabular}{|p{1.5cm}|p{1.6cm}|p{1.6cm}|p{2.2cm}|p{2.6cm}|p{1.8cm}|}%c|c|c|c|}
\hline
Function class   & { \Holder } %&  { \Holder ~ (piecewise smooth)}  
& Besov
&
mixed smooth Besov
%Barron class
& \Holder on a low-dimensional set
% {
%m-Sobolev \phantom{aaa} ($0 < \beta \leq 2$)}
& anisotropic Besov \\
\hline
\hline
Author  & \small \citet{2017arXiv170806633S} 
& \small \citet{suzuki2018adaptivity} %\citet{barron1993universal}
& \citet{suzuki2018adaptivity}
&
\scriptsize \citet{JMLR:v21:20-002,schmidt2019deep,2019arXiv190801842C}
%\multicolumn{1}{|c|}{----}
 &   This work  \\ %[10ex]
\hline 
 Estimation error
& $\Otilde(n^{-\frac{2\beta}{2\beta + d}})$ 
%& $\Otilde(n^{-\frac{1}{2}})$ 
&$\Otilde(n^{-\frac{2\beta}{2\beta +d }})$ %\multicolumn{1}{|c|}{----}
&$ \Otilde\Big(n^{-\frac{2\beta}{2\beta +1}} \times$ {\scriptsize $\log(n)^{\frac{2(d-1)(u+\beta)}{1+2\beta}}\Big) $} %\multicolumn{1}{|c|}{----}
&
$\Otilde(n^{-\frac{2\beta}{2\beta +D}})$
%$n^{- \frac{2s}{2s + 1}}\log(n)^{\frac{2(d-1)(u + s)}{1+2s}} \log(n)^2.
%$
&
 $\Otilde(n^{-\frac{2\betatild}{2\betatild + 1}})$
% $\Otilde(n^{-\frac{2\beta}{2\beta + 1}} \times \log(n)^{\frac{2(d-1)(u + \beta)}{1+2\beta}})$
\\
\hline
\end{tabular}
}
\vspace{-0.3cm}
\end{table*}

From these analyses, one can see that the approximation errors and estimation errors are strongly influenced by two factors, i.e.,
the {\it smoothness} of the target function and the {\it dimensionality} of the input (see Table \ref{tab:RelationBetWork}).
In particular, they suffer from the {\it curse of dimensionality}, which is unavoidable. 
However, these analyses are about the worst case errors and do not exploit specific intrinsic properties of the true distributions.
For example, practically encountered data usually possess low intrinsic dimensionality, i.e., data are distributed on a low dimensional sub-manifold of the input space \citep{tenenbaum2000global,belkin2003laplacian}.
Recently, \citet{JMLR:v21:20-002,schmidt2019deep,chen2019efficient,2019arXiv190801842C} have shown that deep ReLU network has adaptivity to the intrinsic dimensionality of data and can avoid curse of dimensionality if the intrinsic dimensionality is small.
%This analysis strongly supports effectiveness of deep learning on high dimensional settings. 
However, one drawback is that they assumed {\it exact} low dimensionality of the input data.
This could be a strong assumption because practically observed data are always noisy, and injecting noise immediately destroys the low-dimensional structure. Therefore, we consider another direction in this paper.
In terms of curse of dimensionality, \citet{suzuki2018adaptivity} showed that deep learning can alleviate the curse of dimensionality to estimate functions in a so called mixed smooth Besov space (m-Besov). However, m-Besov space assumes strong smoothness toward {\it all} directions uniformly and does not include the ordinary Besov space as a special case. %, while the anisotropic Besov does. 
Moreover, the convergence rate includes heavy poly-log term which is not negligible (see Table \ref{tab:RelationBetWork}).

In practice, one of the typically expected properties of a true function on high-dimensional data is that it is invariant against perturbations of an input in some specific directions (Figure \ref{fig:LowDimManifold}).
For example, in image-recognition tasks, the target function must be invariant against the spatial shift of an input image, which is utilized by data-augmentation techniques \citep{simard2003best,krizhevsky2012imagenet}.
In this paper, we investigate the approximation and estimation abilities of deep learning on {\it anisotropic Besov spaces} 
\citep{nikol1975approximation,Vybiral:2006,triebel2011entropy} (also called dominated mixed-smooth Besov spaces).
An anisotropic Besov space is a set of functions that have ``direction-dependent'' smoothness, whereas ordinary function spaces such as H\"{o}lder, Sobolev, and Besov spaces assume isotropic smoothness that is uniform in all directions. 
We consider a composition of functions included in an anisotropic Besov space, including several existing settings as special cases; it includes analyses of the 
\Holder space \cite{2017arXiv170806633S} and Besov space \cite{suzuki2018adaptivity}, as well as the low-dimensional sub-manifold setting \citep{JMLR:v21:20-002,schmidt2019deep,chen2019efficient,2019arXiv190801842C}\footnote{We would like to remark that the analysis of \cite{JMLR:v21:20-002} does not require smoothness of the embedded manifold that is not covered in this paper.}.
By considering such a space, we can show that %the approximation error and estimation error 
deep learning can alleviate curse of dimensionality if the smoothness in each direction is highly anisotropic. Interestingly, %it is shown that 
any linear estimator (including kernel ridge regression) has worse dependence on the dimensionality than deep learning. 
%Moreover, the adaptivity of deep learning also yields better error than that of linear estimators when the smoothness of the target function is not spatially homogeneous; that is, if there appears a bump or discontinuous shape in the target function, then deep learning is better. 
Our contributions can be 
summarized as follows:
{
\vspace{-0.2cm}
\setlength{\leftmargini}{0.5cm}
\begin{itemize}[itemsep=0.1cm]
\item 
We consider a situation in which the target function is included in a class of anisotropic Besov spaces and show that deep learning can avoid the curse of dimensionality {\it even if the input data do not lie on a low-dimensional manifold}. 
Moreover, deep learning can achieve the optimal adaptive approximation error rate and minimax optimal estimation error rate.
\item 
We compare deep learning with general linear estimators (including kernel methods) and show that deep learning has better dependence on the input dimensionality than linear estimators. 
 %and spatial inhomogeneity of the target function's smoothness 
\end{itemize}
}

\section{Problem setting and the model}

In this section, we describe the problem setting considered in this work. 
We consider the following nonparametric regression model: %For the statistical analysis, we assume the following nonparametric regression model:
$$
\textstyle
y_i = \ftrue(x_i) + \xi_i~~~~(i=1,\dots,n),
$$
where $x_i$ is generated from a probability distribution $P_X$ on $[0,1]^d$,   %, % with density $0 \leq p_X(x) < R$ 
$\xi_i \sim N(0,\sigma^2)$,
and the data $D_n = (x_i,y_i)_{i=1}^n$ are independently identically distributed.
$\ftrue$ is the true function that we want to estimate.
We are interested in the mean squared estimation error of an estimator $\fhat$:
$
\EE_{D_n}[\|\fhat - \ftrue\|_{\LPiPx}^2],
$
where $\EE_{D_n}[\cdot]$ indicates the expectation with respect to the training data $D_n$.
We consider a least-squares estimator in the deep neural network model as $\fhat$ (see \Eqref{eq:ERMprocedure})
and discuss its optimality.
More specifically, we investigate how the ``intrinsic dimensionality'' of data affects the estimation accuracy of deep learning.
For this purpose, we consider an  {\it anisotropic Besov space}  as a model of the target function.
%We will show that the target function varies toward only few directions, then deep learning can avoid the curse of dimensionality. 
%Moreover, we compare deep learning with non-adaptive approximation methods and linear estimators in terms of approximation error and estimation error respectively.

%deep neural network model with non-adaptive methods 
%the estimation error of the deep learning with that of the {\it linear estimator} that includes the kernel ridge regression as a  special case.
%We will show that if the target function is included in the anisotropic Besov space, 
%Actually, we will show that deep learning can outperform {\it any} linear estimators in terms of worst case risk. 
%To derive an estimation error bound, we need to restrict the class of the target functions $\ftrue$.
%In this paper, we consider {\it anisotropic Besov space} as the target function class which we will introduce right below.
%To estimate the true function $\ftrue$ from the data,
%we consider a regularized learning procedure:
%%$\calF(L,W,S,B)$: ReLU-NN with width $W$, depth $L$ ans sparsity $S$ with parameters are bounded by $B$.
%\begin{equation}
%\fhat = \argmin_{f \in \text{set of deep NN}} ~~\sum_{i=1}^n (y_i - f(x_i))^2.
%%\label{eq:ERMprocedure}
%\end{equation}
%where $\bar{f}$ is the {\it clipping} of $f$ defined by $\bar{f} = \min\{\max\{f, - F\},F\}$ for $F > 0$
%which is realized by ReLU units.

%\section{Model: Anisotropic Besov space}

\subsection{Anisotropic Besov space}
In this section, we introduce the anisotropic Besov which was investigated as the model of the true function in this paper.
Throughout this paper, we set the domain of the input to $\Omega = [0,1]^d$. 
%To deal with the output from the internal layer, we let $\Omega^m = [0,1]^m$.
% and $\|g\|_p := (\int_{x \in \Real^d} |g(x)|^p \dd x)^{1/p}$.
For a function $f:\Omega \to \Real$, let $\|f\|_p := \|f\|_{L^p(\Omega)} := (\int_\Omega |f|^p \dd x )^{1/p}$
for $0 < p < \infty$.
For $p=\infty$, we define $\|f\|_\infty :=\|f\|_{L^\infty(\Omega)} := \sup_{x \in \Omega} |f(x)|$.
For $\beta \in \Real_{++}^d$, let $|\beta| = \sum_{j=1}^d |\beta_j|$\footnote{
We let $\Natural := \{1,2,3,\dots\}$, $\Integer_+ := \{0,1,2,3,\dots\}$, $\Integer_+^d := \{(z_1,\dots,z_d) \mid z_i \in \Integer_+ \}$,
$\Real_+ := \{x \geq 0 \mid x \in \Real\}$, and $\Real_{++} := \{x > 0 \mid x \in \Real\}$.
We let $[N] := \{1,\dots,N\}$ for $N \in \Natural$.
}.

For a function $f:\Real^d \to \Real$, we define the $r$th difference of $f$ in the direction $h \in \Real^d$ %at $x \in \Real^d$
as 
\begin{align*}
\Delta_h^r(f)(x)  & :=  % \begin{cases} 
\Delta_h^{r-1}(f)(x+h) - \Delta_h^{r-1}(f)(x),~~ % ~~ \\ %~&  ( x\in \Omega, ~ x+r h\in \Omega), \\ % \sum_{j=0}^r {r \choose j} (-1)^{r-j} f(x+jh)~& 
\Delta_h^0(f)(x)   := f(x),
% ( x\in \Omega, ~ x+r h\in \Omega), \\
%0 & (\text{otherwise}),
%\end{cases}
\end{align*}
for $x \in \Omega$ with $x+r h\in \Omega$, otherwise, let $\Delta_h^r(f)(x) = 0$.
%where $\Delta_h^0(f)(x) = f(x)$.

\begin{Definition}
\label{def:Besov}
For a function $f \in L^p(\Omega)$ where $p \in (0,\infty]$,
the $r$-th modulus of smoothness of $f$ is defined by 
$
w_{r,p}(f,t) = \sup_{h \in \Real^d: |h_i| \leq t_i} \|\Delta_h^r(f)\|_{p},
$
for $t = (t_1,\dots,t_d),~t_i >0$.
\end{Definition}
With this modulus of smoothness, we define the anisotropic Besov space $B^\beta_{p,q}(\Omega)$ for $\beta = (\beta_1,\dots,\beta_d)^\top \in \Real_{++}^d$ as follows.
%where $0 < \beta_1,\dots,\beta_d$ as follows.
\begin{Definition}[Anisotropic Besov space ($B^\beta_{p,q}(\Omega)$)]
For $0 < p,q \leq \infty$, $\beta=(\beta_1,\dots,\beta_d)^\top \in \Real_{++}^d$, 
$r:=  \max_i \lfloor \beta_i \rfloor + 1$, 
let the seminorm $|\cdot|_{B^\alpha_{p,q}}$ be 
\begin{align*}
\textstyle
& |f|_{B^\beta_{p,q}} :=  
%\textstyle
\begin{cases}
%\left(\int_0^\infty (t^{-\alpha} w_{r,p}(f,t))^q \frac{\dd t}{t} \right)^{\frac{1}{q}} & (q < \infty), 
\left( \sum\limits_{k=0}^\infty [2^k w_{r,p}(f, (2^{-k/\beta_1}, \dots, 2^{-k/\beta_d})) ]^q \right)^{1/q} & (q < \infty), \\
\sup_{k \geq 0} 2^k w_{r,p}(f, (2^{-k/\beta_1}, \dots, 2^{-k/\beta_d}))  & (q = \infty).
\end{cases}
\end{align*}
%and 
%$|f|_{B^\alpha_{p,\infty}} := \sup_{t>0} t^{-\alpha} w_{r,p}(f,t) $ for $q=\infty$.
The norm of the anisotropic Besov space $B_{p,q}^\beta(\Omega)$ is defined by 
$\|f\|_{B_{p,q}^\beta} := \|f\|_{p} + |f|_{B^\beta_{p,q}}$,
and %we have 
$B^\beta_{p,q}(\Omega) = \{f \in L^p(\Omega) \mid \|f\|_{B_{p,q}^\beta} < \infty\}$.
\end{Definition}

Roughly speaking $\beta$ represents the smoothness in each direction.
If $\beta_i$ is large, then a function in $B_{p,q}^\beta$ is smooth to the $i$th coordinate direction,
otherwise, it is non-smooth to that direction.
$p$ is also an important quantity that controls the {\it spatial inhomogeneity} of the smoothness.
%For small $p$ (i.e., $p=1$), the smoothness is ensured only in an average sense with respect to input $x$, but 
%for large $p$ (i.e., $p=\infty$), the smoothness is ensured uniformly (see Proposition \ref{lemm:EmbeddingLemma} for more details).
If $\beta_1 = \beta_2 = \dots = \beta_d$, then the definition is equivalent to the usual Besov space \citep{devore1988interpolation,devore1993wavelet}.
\citet{suzuki2018adaptivity} analyzed curse of dimensionality of deep learning through a so-called {\it mixed smooth Besov} (m-Besov) space which imposes a stronger condition toward all directions uniformly. Particularly, it imposes stronger smoothness toward non-coordinate axis directions. 
Moreover, m-Besov space does {\it not} include the vanilla Besov space as a special case and thus cannot capture the situation that we consider in this paper.

Throughout this paper, for given $\beta = (\beta_1,\dots,\beta_d)^\top \in \Real_{++}^d$, we write 
$\betamin := \min_i \beta_i$ (smallest smoothness) and $\betamax := \max_i \beta_i$ (largest smoothness). 
The approximation error of a function in anisotropic Besov spaces is characterized by the harmonic mean of $(\beta_j)_{j=1}^d$, which corresponds to the average smoothness, and thus we define 
\begin{align}
\textstyle 
\betatild := %\left( \sum_{j=1}^d \frac{1}{\beta_j}\right)^{-1}.
\left( \sum_{j=1}^d 1/\beta_j \right)^{-1}.
\label{eq:betatilddef}
\end{align}
The Besov space is closely related to other function spaces such as \Holder space.
Let $
\partial^\alpha f(x) %= \frac{\partial f}{\partial^\alpha x}(x) 
= \frac{\partial^{|\alpha|} f}{\partial^{\alpha_1} x_1 \dots  \partial^{\alpha_d} x_d}(x).
$

\begin{Definition}[\Holder space ($\calC^\beta(\Omega)$)]
For a smoothness paraemter $\beta \in \Real_{++}$ with $\beta \not \in \Natural$,
consider an $m$ times differentiable function $f:\Real^d \to \Real$
where $m=\lfloor \beta \rfloor$ (the largest integer less than $\beta$),
and let the norm of the \Holder space $\calC^\beta(\Omega)$ be 
$
%\calC^\beta_d(K) = 
%\left\{f % :[0,1]^d\to \Real
%~\big |~ 
\|f\|_{\calC^\beta} := \max_{|\alpha| \leq m}
\big\|\partial^\alpha f\|_{\infty}
+ \max_{|\alpha| = m} \sup_{x,y \in\Omega}
\frac{|\partial^\alpha f(x) - \partial^\alpha f(y)|}{\|x-y\|^{\beta -m}}. %\leq K \right\}
$
Then,  ($\beta$-)\Holder space $\calC^\beta(\Omega)$ is defined as %a set of functions $f$ with finite norm 
$\calC^\beta(\Omega) = \{f \mid \|f\|_{\calC^\beta} < \infty\}$.
\end{Definition}
Let $\calC^0(\Omega)$ be the set of continuous functions equipped with $L^\infty$-norm:
$\calC^0(\Omega) := \{ f : \Omega \to \Real \mid \text{$f$ is continuous and $\|f\|_\infty < \infty$} \}$.
%Then, 
These function spaces %introduced above 
are closely related to each other.  %\citep{triebel2011entropy}.
\begin{Proposition}[\citet{triebel2011entropy}]
\label{lemm:EmbeddingLemma}
There exist the following relations between the spaces:
\vspace{-0.1cm}
{\setlength{\leftmargini}{20pt}  
\begin{enumerate}
\setlength{\itemsep}{-5pt}
\item For $\beta =(\beta_0,\dots,\beta_0)^\top \in \Real^d$ with $\beta_0 \not \in \mathbb{N}$, it holds that
%\begin{equation}
$\calC^{\beta_0}(\Omega) = B^\beta_{\infty,\infty}(\Omega).$
%\label{eq:HolderEquivalence}
%\end{equation}
%\item For $0 < s <m < \infty$, $1 \leq p < \infty$, $q \leq \infty$, 
%$B^s_{p,q}(\Omega) = [L^p(\Omega),W_p^m(\Omega)]_{{s/m},q}.$
%\item For $0 < s_1 < s < s_2$, $s=(1-\theta) s_1 + \theta s_2$ and $1 \leq q_1,q_2 \leq \infty$, 
%$B^s_{p,q}(\Omega) = [B^{s_1}_{p,q_1}(\Omega),B^{s_2}_{p,q_2}(\Omega)]_{\theta,q}.$
\item For $0 < p_1,p_2,q \leq \infty$, $p_1 \leq p_2$ and $\beta \in \Real_{++}^d$ with 
$\betatild > (1/p_1 - 1/p_2)_+$\footnote{Here, we let $(x)_+ := \max\{x,0\}$.}, it holds that\footnote{The symbol $\hookrightarrow$ means continuous embedding.} 
%\begin{equation}
$B_{p_1,q}^\beta(\Omega) \hookrightarrow B_{p_2,q}^{\gamma \beta}(\Omega)$
for $\gamma =  1- (1/p_1 - 1/p_2)_+/\betatild$.
%\label{eq:BpPembed}
%\end{equation}
\item For $0 < p,q_1,q_2 \leq \infty$, $q_1 < q_2$, and $\beta \in \Real_{++}^d$, it holds that 
$
B_{p,q_1}^{\beta} \hookrightarrow B_{p,q_2}^{\beta}
$. In particular, %combined 
with properties 1 and 2, %with \Eqref{eq:HolderEquivalence} and \Eqref{eq:BpPembed}, 
if $\betatild > 1/p$, it holds that 
%\begin{equation}
%\label{eq:BesovHolderEmbed}
$B_{p,q}^{\beta}(\Omega)  \hookrightarrow \calC^{\gamma \betamin}(\Omega)$ 
%\end{equation}
where $\gamma = 1 - 1/(\betatild p)$.
%In particular, under the same condition, from the definition of $\|\cdot\|_{B_{p,q}^s}$, it holds that
%\begin{align}
%%\textstyle
%B_{p,q}^s(\Omega) \hookrightarrow L^r(\Omega).
%\label{eq:BesovLrEmbedding}
%\end{align}
% \cite{cohen2001tree} %% L^r embedding is proven, but the more general form is not given.
\item For $0 < p,q \leq \infty$ and $\beta \in \Real_{++}^d$, if $\betatild > 1/p$, then 
%\begin{align}
$B_{p,q}^\beta(\Omega) \hookrightarrow \calC^0(\Omega)$.
%\label{eq:BesovContEmbedding}
%\end{align}
\end{enumerate}
}
\end{Proposition}
This result is basically proven by \citet{triebel2011entropy}. 
%For completeness, we provide its derivation in Appendix \ref{sec:EmbeddingTheorem} in the supplementary material.
%If the average smoothness $\betatild$ is sufficiently large ($\betatild > 1/p$), then the functions in $B_{p,q}^\beta$ are continuous;
%but if it is small ($\betatild < 1/p$), then they are no longer continuous.
%Actually, small $p$ indicates spatially inhomogeneous smoothness, and thus there would appear spikes and jumps (see \cite{donoho1998minimax} for this perspective from a view point of wavelet analysis).
%
For completeness, we provide its derivation in Appendix \ref{sec:EmbeddingTheorem}.
If the average smoothness $\betatild$ is sufficiently large ($\betatild > 1/p$), then the functions in $B_{p,q}^\beta$ are continuous;
however, if it is small ($\betatild < 1/p$), then they are no longer continuous.
Small $p$ indicates spatially inhomogeneous smoothness; thus, spikes and jumps appear (see \citet{donoho1998minimax} for this perspective, from the viewpoint of wavelet analysis).

%The embedding property (property 3) %\eqref{eq:BesovHolderEmbed} 
%is useful to analyze the deep composition model, which will be introduced in the following sections,
%because the continuity of functions controls how strongly the approximation error in internal layers can propagate to the last layer.

\subsection{Model of the true function}
\label{sec:ModelTrueFunction}
As a model of the true function $\ftrue$, we consider two types of models:
\emph{Affien composition model} and \emph{deep composition model}.
For a Banach space $\calH$, we let $U(\calH)$ be the unit ball of $\calH$.

%\paragraph{Affine composition model}
{\bf (a) Affine composition model:}
The first model we introduced is a very naive model which is just a composition of 
an affine transformation and a function in the anisotropic Besov space:
\begin{align*}
& \calHaff := \{  h(A x + b) \mid 
h \in U(B^{\beta}_{p,q}([0,1]^{\dtil})), 
%\\  &
~A \in \Real^{\dtil \times d},~b \in \Real^b~
\text{s.t.}~
A x + b \in [0,1]^{\dtil}~(\forall x\in \Omega)
\},
\end{align*}
%where $h \in B^{\beta}_{p,q}([0,1]^m)$, $A \in \Real^{m \times d}$ and $b \in \Real^b$.
where we assume $\dtil \leq d$.
Here, we assumed that the affine transformation has an appropriate scaling such that  
$A x + b$ is included in the domain of $h$ for all $x \in \Omega$.
This is a quite naive model but  provides an instructive example to understand how the estimation error of deep learning behaves under the anisotropic setting.

%\paragraph{Deep composition model}
%~
{\bf (b) Deep composition model:}
The {\it deep composition model} generalizes the affine composition model to a composition of nonlinear functions. % and covers several important examples.
Let $\melle{1} = d$, $\melle{L+1} = 1$, $\mell$ be the dimension of the $\ell$th layer, 
and let $\beta^{(\ell)} \in \Real_{++}^{\mell}$ be the smoothness parameter in the $\ell$th layer.
The deep composition model is defined as 
\begin{align*}
& \calHdeep := \{  h_H \circ \dots \circ h_1(x) \mid 
 h_\ell: [0,1]^{\mell} \!\! \to \! [0,1]^{\melle{\ell+1}},~% \\ & 
h_{\ell,k} \in U(B_{p,q}^{\beta^{(\ell)}}([0,1]^{\mell}))~(\forall k \in [\melle{\ell+1}])\}.
\end{align*}
%where $h_\ell: [0,1]^{\mell} \to [0,1]^{\melle{\ell+1}}$.
%Let $h_{\ell,k}$ be the $k$-th output of $h_\ell$ (i.e., $h_\ell(x) = (h_{\ell,1}(x),\dots,h_{\ell,\melle{\ell + 1}}(x))^\top \in \Real^{\melle{\ell+1}}$),
%then we assume that $h_{\ell,k} \in h_\ell: [0,1]^{\mell} \to [0,1]^{\melle{\ell+1}}$.
Here, the interval $[0,1]$ can be replaced by another compact interval, such as $[a_{\ell},b_\ell]$, but this difference can be absorbed by changing a scaling factor. The assumption $\|h_{\ell,k}\|_{B_{p,q}^{\beta^{(\ell)}}} \leq 1$ can also be relaxed, but we do not pursue that direction due to presentation simplicity.
This model includes the affine composition model as a special case. 
However, it requires a stronger assumption to properly evaluate the estimation error on this model. 

%\subsubsection{Examples}
\paragraph{Examples}
The model we have introduced includes some instructive examples as listed below: 
%that have been investigated.

{\bf  (a) Linear projection}~~
\citet{2017arXiv170806633S} analyzed estimation of the following model by deep learning:
$
\ftrue(x) = g(w^\top x)
$
where $g \in \calC^{\beta}([0,1])$ and $w \in \Real^d$.
In this example, the function $\ftrue$ varies along only one direction, $w$.
Apparently, this is an example of the affine composition model.
%This model has another representation $h(Ux)$, where $U$ is a rotation matrix such that $U w$ is parallel to $(1,0,\dots,0)^\top$,
%and $h$ is an element of $B^{\alpha}_{\infty,\infty}$ with $\alpha = (\beta,\gamma,\dots,\gamma)^\top \in \Real_{++}^d$, where $\gamma > 0$ can be arbitrary large.
%The intuition behind this is that after rotation $U$, the function depends only one direction $(1,0,\dots,0)^\top$, and thus it can be arbitrarily smooth toward other directions perpendicular to the first coordinate. 
%Our formulation of the affine composition model can be seen as an extension of this example.

\begin{figure}[tb]
%\begin{wrapfigure}[17]{r}{70mm}
%\vspace*{-\intextsep}
\begin{minipage}[t]{0.55\linewidth} 
\centering
\includegraphics[width=6.5cm,clip,pagebox=artbox]{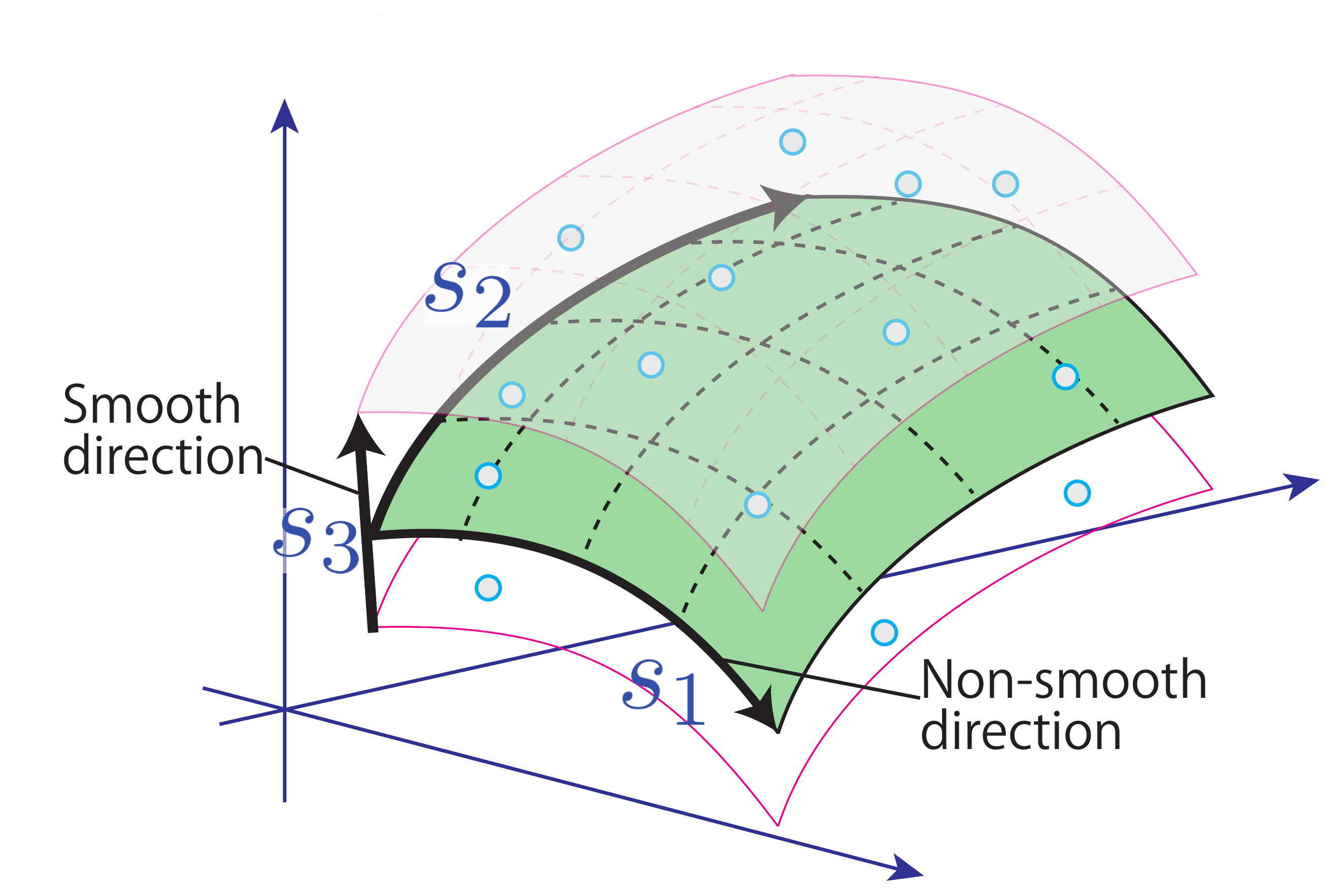}%5.3cm,clip,pagebox=artbox]{./fig/low_dim_manifold_with_coordinate.pdf}
\end{minipage}
\begin{minipage}[b]{0.35\linewidth} 
\captionof{figure}{Near low dimensional data distribution with anisotropic smoothness of the target function.
The target function has less smoothness ($s_1,s_2$) toward the first two coordinates on the manifold while it is almost constant toward the third coordinate (large $s_3$).}
\label{fig:LowDimManifold}
\end{minipage}
\vspace{-0.3cm}
\end{figure}
%\end{wrapfigure}

{\bf  (b) Distribution on low dimensional smooth manifold}~~
Assume that the input $x$ is distributed on a low-dimensional smooth manifold embedded in $\Omega$,
and the smoothness of the true function $\ftrue$ is anisotropic along a coordinate direction on the manifold.
We suppose that the low dimensional manifold is $\dtil$-dimensional and $\dtil \ll d$.
In this situation, the true function can be written as 
$
\ftrue(x) = h(\phi(x))
$
where $\phi:\Real^d \to \Real^{\dtil}$ is a map that returns the coordinate of $x$ on the manifold and 
$h$ is an element in an anisotropic Besov space on $\Real^{\dtil}$.
This %kind of 
situation appears if data is distributed on a low-dimensional sub-manifold of $\Omega$ and the target function is invariant against noise injection to some direction on the manifold at each input point $x$
(Figure \ref{fig:LowDimManifold} illustrates this situation).
One typical example of this situation is a function invariant with data augmentation \citep{simard2003best,krizhevsky2012imagenet}. Even if the noise injection destroys low dimensionality of the data distribution (i.e., $\dtil = d$), an anisotropic smoothness of the target function eases the curse of dimensionality as analyzed below, which is quite different from existing works \citep{yang2016bayesian,bickel2007local,JMLR:v21:20-002,schmidt2019deep,chen2019efficient,2019arXiv190801842C}.

%\subsection{Related work}
\paragraph{Related work}
Here, we introduce some more related work and discuss their relation to our analysis.
%\cite{ibragimov1984more,nyssbaum1987nonparametric}.
The statistical analysis on an anisotropic Besov space can be back to \citet{ibragimov1984more} who considered density estimation, where the density is assumed to be included in an anisotropic Sobolev space with $p \geq 2$, and derived the minimax optimal rate $n^{-r \betatild/(2\betatild + 1)}$ with respect to $L^r$-norm.
\citet{nyssbaum1983optimalfiltration,nyssbaum1987nonparametric} analyzed a nonparametric regression problem on an anisotropic Besov space.
Following these results, several studied have been conducted %More studies have been devoted for estimation problem in the anisotropic Besov 
in the literature pertaining to nonparametric statistics, such as 
nonlinear kernel estimator \cite{Kerkyacharian2001}, adaptive confidence band construction \cite{hoffman2002}, optimal aggregation \cite{gaiffas2011hyper}, Gaussian process estimator \cite{bhattacharya2011adaptive,bhattacharya2014anisotropic}, and kernel ridge regression \cite{hang2018optimal}.
%\cite{Kerkyacharian2001} analyzed a nonlinear kernel estimator for the white-noise denoising problem, where the true function is in an anisotropic Besov space.
%\cite{hoffman2002} proposed an estimator that is adaptive to the unknown smoothness $\beta$ and derived an adaptive confidence band.
%\cite{gaiffas2011hyper} investigated an optimal aggregation procedure that adapts to the unknown smoothness parameter $\beta$.
%\cite{bhattacharya2014anisotropic} proposed a Gaussian process estimator that is also adaptive to the smoothness parameter $\beta$.
%\cite{bhattacharya2011adaptive} investigated a Gaussian process estimator with dimension reduction on function spaces with anisotropic smoothness. 
%\cite{hang2018optimal} investigated a kernel ridge regression-type estimator on an anisotropic Besov space $B^\beta_{p,\infty}$ with $p \geq 2$,
%and proved its minimax optimality.
Basically, these studies investigated estimation problems in which the target function is in anisotropic Besov spaces, but the composition models considered in this paper have not been analyzed.
\citet{hoffman2002,bhattacharya2011adaptive} considered a dimension reduction model; that is, the target function is dependent on only a few variables of $x$, but they did not deal with more general models, such as the affine/deep composition models.
%By considering such composition models, we can show sub-optimality of the linear estimators (see Section \ref{sec:MinimaxRateOfLinearEstimator}
%), whereas the deep neural network model is suitable to estimate composition models.
The nonparametric regression problems where the input data are distributed on a low-dimensional smooth manifold has been studied as a ``manifold regression'' \cite{yang2016bayesian,bickel2007local,yang2015minimax}.
Such a model can be considered as a specific example of the deep composition model.
In this sense, our analysis is a significant extension of these analyses.
%However, our analysis is not specified to this model, and thus the dependence of the constant coefficient hidden in the $O(\cdot)$-symbol is not optimized with respect to the input dimensionality $d$ and the manifold dimensionality $\dtil$.
%We believe it is easy to improve this point, but we do not pursue this direction in this study. 

\section{Approximation error analysis}
\label{sec:ApproximatoinErroBounds}

Here, we consider approximating the true function $\ftrue$ via a deep neural network
and derive the approximation error.
As the activation function, we consider the ReLU activation denoted by $\eta(x) = \max\{x,0\}~(x \in \Real)$.
For a vector $x$, $\eta(x)$ is operated in an element-wise manner.
The model of neural networks with height $L$, width $W$, sparsity constraint $S$, and norm constraint $B$ as 
$
 \Phi(L,W,S,B)
:= \{ (\Well{L} \eta( \cdot) + \bell{L}) \circ \dots  
\circ (\Well{1} x + \bell{1})
\mid 
 \Well{L} \in \Real^{1 \times W},~\bell{L} \in \Real,~  
\Well{1} \in \Real^{W \times d},~\bell{1} \in \Real^W, 
 \Well{\ell} \in \Real^{W \times W},~\bell{\ell} \in \Real^W (1 < \ell < L), 
%\textstyle 
\sum\nolimits_{\ell=1}^L (\|\Well{\ell}\|_0+ \|\bell{\ell}\|_0) \leq S, 
\max_{\ell} \|\Well{\ell}\|_\infty \vee \|\bell{\ell}\|_\infty \leq B
\},
$
%\begin{align*}
%& \Phi(L,W,S,B) \\ & 
%:= \{ (\Well{L} \eta( \cdot) + \bell{L}) \circ \dots  
%\circ (\Well{1} x + \bell{1})
%\mid \\  
%& \Well{L} \in \Real^{1 \times W},~\bell{L} \in \Real,~  
%\Well{1} \in \Real^{W \times d},~\bell{1} \in \Real^W, \\
%& \Well{\ell} \in \Real^{W \times W},~\bell{\ell} \in \Real^W (1 < \ell < L), \\ & 
%%\textstyle 
%\sum\nolimits_{\ell=1}^L (\|\Well{\ell}\|_0+ \|\bell{\ell}\|_0) \leq S, \\ &
%\max_{\ell} \|\Well{\ell}\|_\infty \vee \|\bell{\ell}\|_\infty \leq B
%\},
%\end{align*}
where $\|\cdot\|_0$ is the $\ell_0$-norm of the matrix (the number of non-zero elements of the matrix),
and $\|\cdot\|_\infty$ is the $\ell_\infty$-norm of the matrix (maximum of the absolute values of the elements).
The sparsity constraint and norm bounds are required to obtain the near-optimal rate of the estimation error. 
To evaluate the accuracy of the deep neural network model in approximating target functions, 
we define the worst-case approximation error as
$$
\textstyle
R_r(\calF,\calH) :=\sup_{f^* \in \calH}  \inf_{f \in \calF} \|f^* - f\|_{L^r(\Omega)},
$$
where $\calF$ is the set of functions used for approximation, and $\calH$ is the set of target functions. % that have to be approximated.

\begin{Proposition}[Approximation ability for anisotropic Besov space]
\label{prop:BesovApproxByNN}
Suppose that $0 < p,q,r \leq \infty$ and $\beta \in \Real_{++}^d$ satisfy
the following condition: %s:
%\begin{align}
%\textstyle
$\betatild > (1/p - 1/r)_+$.
%\label{eq:sprConditions}
%\\~~\text{(ii)} & ~~~   s = d/p, q \leq \min\{1,r\}, p,r < \infty.& & &
%\end{align}
Assume that $m \in \Natural$ satisfies $0 < \betamax < \min( m , m -1 + 1/p)$.
%\end{subequations}
%\begin{subequations}
%\label{eq:sprConditions}
%\begin{flalign}
%~~\text{(i)} &  ~~~  s > d(1/p - 1/r)_+, &  & & 
%\\~~\text{(ii)} & ~~~   s = d/p, q \leq \min\{1,r\}, p,r < \infty.& & &
%\end{flalign}
%\end{subequations}
%\begin{enumerate}
%\item[(i)] $s > d(1/p - 1/r)_+$,
%\item[(ii)] $s = d/p$, $q \leq \min\{1,r\}$, $p,r < \infty$.
%\item[(iii)] $r \leq p$.
%\end{enumerate}
Let $\delta = (1/p - 1/r)_+$, $\nu = (\betatild - \delta)/(2\delta)$ and $W_0(d) := 6 d m(m+2) + 2 d$.
Then, for $N \in \Natural$, we can bound the approximation error as  
$$
%\inf_{\fcheck \in \calF(L,W,S,B)} \sup_{\ftrue \in U(B^s_{p,q}([0,1]^d))}\|\ftrue -  \fcheck\|_{L^r([0,1]^d)}
\textstyle
R_r(\Phi(L_1,W_1,S_1,B_1),U(B^\beta_{p,q}(\Omega)))
\lesssim  N^{ - \betatild},
$$
by setting 
\begin{flalign}
%\qquad 
\textstyle & L_1(d) \textstyle := 3 +  2 \lceil \log_2\left(  \frac{3^{d\vee m} }{\epsilon c_{(d,m)}}  \right)+5 \rceil \lceil\log_2(d \vee m)\rceil,~~ \textstyle W_1(d)  := N W_0 ,   \\
\textstyle & S_1(d)  \textstyle := [(L-1) W_0^2  + 1] N,~~ B_1(d) \textstyle := O(N^{d (1 + \nu^{-1})(1/p-\betatild)_+}), 
\label{eq:L1B1def}
\end{flalign}
for $\epsilon = N^{-\betatild} \log(N)^{-1}$ and a constant $c_{(d,m)}$ depending only on $d$ and $m$.
%
%there exists a constant $c_{(d,m)}$ depending only on $d$ and $m$ such that 
%for sufficiently large $N \in \Natural$ and $\epsilon = N^{-\betatild} \log(N)^{-1}$, 
%by setting
%\begin{flalign*}
%\qquad L_1(d) &:= 3 +  2 \lceil \log_2\left(  \frac{3^{d\vee m} }{\epsilon c_{(d,m)}}  \right)+5 \rceil \lceil\log_2(d \vee m)\rceil, \\
%W_1(d) &:= N W_0 , ~~~S_1(d) := [(L-1) W_0^2  + 1] N, \\
% B_1(d) & := O(N^{d (1 + \nu^{-1})(1/p-\betatild)_+}), 
%\end{flalign*}
%\begin{enumerate}
%\item $L = 3 +  2 \lceil \log_2\left(  \frac{3^{d\vee m} }{\epsilon c_{(d,m)}}  \right)+5 \rceil \lceil\log_2(d \vee m)\rceil$, 
%\item $W = N W_0 $,
%\item $S = (L-1) W_0^2 N + N$,
%\item $B = O(N^{(\nu^{-1} + d^{-1}) (1\vee (d/p - s)_+) })$,
%\end{enumerate}
%$\calF(L,W,S,B)$ satisfies 
%Otherwise, it holds that
%On the other hand, even if the condition \eqref{eq:sprConditions} is not satisfied,
%for $0 < p, q, r \leq \infty$ and $0 < s$, it holds that  
%$$
%R_r(\calF(L,W,S,B),B^s_{p,q}([0,1]^d))
%\lesssim  N^{ - s/d + (1/p - 1/r)_+}.
%$$
\end{Proposition}
The proof of this proposition is provided in Appendix \ref{sec:ProofPropBesovApp}.
The rate $N^{-\betatild}$ is the optimal {\it adaptive} approximation error rate that can be achieved by a model with $N$ parameters (the difference between adaptive and non-adaptive methods is explained in the discussion below).
% that can be accomplished by a linear combination of $N$-term basis functions. 
Note that this is an approximation error in an oracle setting and no sample complexity appears here.
We notice that we can avoid the {\it curse of dimensionality} if the average smoothness $\betatild$ is small.
This means that if the target function is non-smooth in only a few directions and smooth in other directions, we can avoid the curse of dimensionality.
In contrast, if we consider an isotropic Besov space where $\beta_1 = \cdots = \beta_d (= \betamin)$, then $\betatild = \betamin/d$, which directly depends on the dimensionality $d$, and we need an exponentially large number of parameters in this situation to achieve $\epsilon$-accuracy.
Therefore, the anisotropic smoothness has a significant impact on the approximation error rate.
The assumption $\betatild > (1/p - 1/r)_{+}$ ensures the $L_r$-integrability of the target function, and the inequality (without equality) admits a near-optimal wavelet approximation of the target function in terms of $L_r$-norm.

Using this evaluation as a basic tool, we can obtain the approximation error for the deep composition models.
We can also obtain the approximation error for the affine composition models, but it is almost identical to Proposition \ref{prop:BesovApproxByNN}. Therefore, we defer it to Appendix \ref{sec:AffineCompApproxError}. % in the supplementary material.

\begin{Theorem}[Deep composition model]
\label{thm:DeepApproximationError}
Assume that 
%\begin{equation}
%\label{eq:BetaEllLinfAss}
%\textstyle
$\betatild^{(\ell)} > 1/p$
%\end{equation}
for all $\ell = 1,\dots,H$.
Then, the estimation error on the deep composition model is bounded as
\begin{equation}
R_\infty(\Phi(L,W,S,B),\calHdeep) \lesssim 
\max_{\ell \in [H]} N^{-\betatild^{*(\ell)}},
\label{eq:DeepCompApprox}
\end{equation}
where $\betatild^{*(\ell)} = \betatild^{(\ell)} \prod_{k=\ell+1}^H [(\betamin^{(k)} -1/p)\wedge 1]$,
and 
$L = \sum_{\ell=1}^H (L_1(\mell) + 1), W = \max_\ell (W_1(\mell) \vee \melle{\ell+1}), S = \sum_{\ell=1}^H (S_1(\mell) + 3 \melle{\ell+1}),
B = \max_\ell B_1(\mell).$
%$L,W,S,B$ are specified in the supplementary material (Appendix \ref{sec:ProofOfAffineDeepApprox}).

\end{Theorem}
The proof can be found in Appendix \ref{sec:ProofOfAffineDeepApprox}.
Since the model is more general than the vanilla anisotropic Besov space,
we require a stronger assumption $\betatild^{(\ell)} > 1/p$ %\eqref{eq:BetaEllLinfAss} 
on $\betatild^{(\ell)}$
than the condition in Proposition \ref{prop:BesovApproxByNN}. %\eqref{eq:sprConditions}.
This is because we need to bound the \Holder smoothness of the remaining layers to bound the influence of the approximation error in the internal layers to the entire function.
\Holder smoothness is ensured according to the embedding property under this condition %\eqref{eq:BetaEllLinfAss}
 (Proposition \ref{lemm:EmbeddingLemma}).
This \Holder smoothness assumption affects the approximation error rate. 
The convergence rate $\betatild^{*(\ell)}$ in \Eqref{eq:DeepCompApprox} is different from that in \Eqref{eq:AffineCompApprox}. 
This is because the approximation error in the internal layers are propagated through
the remaining layers with \Holder smoothness and its amplitude is controlled by the \Holder smoothness.

%\subsection{Approximation error by non-adaptive method}
%\label{sec:ApproxNonAdaptive}
%
%\linepenalty
\vspace{-0.3cm}
\paragraph{Approximation error by non-adaptive method}\looseness=-1
The approximation error obtained in the previous section is called an adaptive error rate in the literature regarding approximation theory \citep{DeVore1998nonlinearapprox}.
If we fix $N$ bases beforehand and approximate the target function by a linear combination of the $N$ bases (which is called the non-adaptive method),
then we {\it cannot} achieve the adaptive error rate obtained in the previous section.
%The error rate for non-adaptive methods can be characterized by the {\it Kolmogorov width} ($N$-term approximation error) \citep{KolWidth:Kolmogoroff:1936,tikhomirov1960diameters}.
Roughly speaking, the approximation error of non-adaptive methods is lower bounded by $N^{-\left(\betatild - (\frac{1}{p} - \frac{1}{\min\{2,r\}})_+\right)}$
 \citep{myronyuk2015trigonometric,Myronyuk2016,Myronyuk2017},
%Therefore, 
%We can see that the approximation error of deep neural networks achieves faster rate especially for small $p$.
which is slower than the approximation error rate of deep neural networks especially for small $p$.

\vspace{-0.1cm}
\section{Estimation error analysis}
\vspace{-0.1cm}

In this section, we analyze the accuracy of deep learning in estimating a function in compositions of anisotropic Besov spaces.
We consider a least-squares estimator in the deep neural network model:
\begin{equation}
\textstyle
\fhat = \argmin_{\bar{f}: f \in \Phi(L,W,S,B)} \sum_{i=1}^n (y_i - \bar{f}(x_i))^2
\label{eq:ERMprocedure}
\end{equation}
where $\bar{f}$ is the {\it clipping} of $f$ defined by $\bar{f} = \min\{\max\{f, - F\},F\}$ for a constant $F > 0$
which is realized by ReLU units.
The network parameters $(L,W,S,B)$ should be specified appropriately as indicated in Theorems \ref{thm:EstimationErrorNNBesov_affine}
and \ref{thm:EstimationErrorNNBesov_Deep}. In practice, these parameters can be specified by cross validation. Indeed, we can theoretically show that
cross validation can provide the appropriate choice of these parameters in compensation to an additional $\log(n)$-factor in the estimation error bound. %; however, we omit such an analysis for simplicity.
This estimator can be seen as a sparsely regularized estimator because there are constraints on $S$.
In terms of optimization, this requires a combinatorial optimization, but we do not pursue the computational aspect.
The estimation error that we derive in this section can involve the optimization error, 
but for simplicity, we only demonstrate the estimation error of the {\it ideal} situation where there is no optimization error.
%We assume the optimal solution can be obtained and obtain the .

\vspace{-0.3cm}
\paragraph{Affine composition model}
The following theorem provides an upper bound of the estimation error for the affine composition model.
\begin{Theorem}
\label{thm:EstimationErrorNNBesov_affine}
Assume the same condition as in Theorem \ref{thm:AffineApproximationError};
in particular, suppose $0 < p,q \leq \infty$ and $\betatild > (1/p - 1/2)_+$ for $\betatild \in \Real_{++}^{\dtil}$.
Moreover, we assume that the distribution $P_X$ has a density $p_X$ such that $\|p_X\|_\infty \leq R$ for a constant $R > 0$.
If $\ftrue \in \calHaff \cap L^\infty(\Omega)$, %B^{\beta}_{p,q}(\Omega) \cap L^\infty(\Omega)$ 
%$h$ $\|h\|_{B^{\beta}_{p,q}} \leq 1$
and $\|\ftrue\|_\infty \leq F$ for $F \geq 1$;  
then, %by letting $N \asymp n^{\frac{1}{1+2s}}$, 
letting $(W,L,S,B)=(L_1(\dtil),W_1(\dtil),S_1(\dtil),(\dtil C +1)B_1(\dtil))$ as in Theorem \ref{thm:AffineApproximationError} with 
$N \asymp n^{\frac{1}{2\betatild+1}}$, 
we obtain 
\begin{align*}
\EE_{D_n}[\|\ftrue -  \fhat\|_{\LPi(P_X)}^2] \lesssim 
n^{- \frac{2\betatild}{2\betatild + 1}} \log(n)^{3},
\end{align*}
where $\EE_{D_n}[\cdot]$ indicates the expectation with respect to the training data $D_n$.
%where $s' = s - 1/p + 1/2$.
\end{Theorem}
%The proof is given in Appendix \ref{sec:ProofsOfEstimationErrorBounds}.
The proof is provided in Appendix \ref{sec:EstimationErrorAnalProof}.
%This is a direct extension of \citet{suzuki2018adaptivity} and \citet{2017arXiv170806633S}, in which an ordinary (isotropic) Besov space and \Holder space were investigated, respectively.
We will show that the convergence rate $n^{-2\betatild/(2\betatild + 1)}$ is minimax optimal in Section \ref{sec:MinimaxOpt}
(see also \citet{Kerkyacharian1992densitybesov,donoho1996densitywaveletminimax,donoho1998minimax,GineNickl2015mathematical} for ordinary Besov spaces). 
%Thus, it cannot be improved by any estimator. Therefore, deep learning can achieve the minimax optimal rate up to $\log(n)^3$-order.
The $L^\infty$-norm constraint $\|\ftrue\|_\infty \leq F$ is used to derive a uniform bound on the discrepancy between 
the population and %$L^2$-norm $\|\fhat - \ftrue\|_{\LPi(P_X)}^2$ and 
the empirical $L^2$-norm. % $\frac{1}{n} \sum_{i=1}^n (\fhat(x_i) - \ftrue(x_i))^2$.
Without this condition, the convergence rate could be slower.
%If we consider a fixed design risk, $\EE_{D_n}[\frac{1}{n} \sum_{i=1}^n (\fhat(x_i) - \ftrue(x_i))^2]$,
%then this assumption can be removed. 
%The term $\log(n)^3$ could be improved to $\log(n)^2$ by using the construction of \cite{petersen2017optimal}.
%However, we don't pursue this direction for simplicity.  

\vspace{-0.3cm}
\paragraph{Deep composition model}
For the deep composition model, we obtain the following convergence rate. This is an extension of Theorem \ref{thm:EstimationErrorNNBesov_affine} but requires a stronger assumption on the smoothness. % parameter.
\begin{Theorem}
\label{thm:EstimationErrorNNBesov_Deep}
Suppose that $0 < p,q \leq \infty$ and $\betatild^{(\ell)}> 1/p$ for all $\ell \in [H]$.
If $\ftrue \in \calHdeep \cap L^\infty(\Omega)$, %B^{\beta}_{p,q}(\Omega) \cap L^\infty(\Omega)$ 
%$h$ $\|h\|_{B^{\beta}_{p,q}} \leq 1$
and $\|f\|_\infty \leq F$ for $F \geq 1$,  
then %by letting $N \asymp n^{\frac{1}{1+2s}}$, 
%letting $(W,L,S,B)$ be as in Proposition \ref{prop:BesovApproxByNN} with 
%$N \asymp n^{\frac{1}{2\betatild+1}}$, 
we obtain 
\begin{align*}\textstyle
\EE_{D_n}[\|\ftrue -  \fhat\|_{\LPi(P_X)}^2] \lesssim 
\max_{\ell \in [H]}
n^{- 2\betatild^{*(\ell)}/(2\betatild^{*(\ell)} + 1)} \log(n)^{3},
\end{align*}
where $\betatild^{*(\ell)}$ is defined in Theorem \ref{thm:DeepApproximationError},
and $(L,W,S,B)$ is as given in Theorem \ref{thm:DeepApproximationError} with $N \asymp \max_{\ell \in [L]} n^{\frac{1}{2\betatild^{*(\ell)} +1}}$.
%where $\EE_{D_n}[\cdot]$ indicates the expectation w.r.t. the training data $D_n$.
%where $s' = s - 1/p + 1/2$.
\end{Theorem}

The proof is provided in Appendix \ref{sec:EstimationErrorAnalProof}.
We will show that this is also minimax optimal in Theorem \ref{thm:MinimaxRate}.
Because of the \Holder continuity, the convergence rate becomes slower than the affine composition model (that is, $\betatild^{*(\ell)} \leq \betatild^{(\ell)}$). However, this slower rate is unavoidable in terms of the minimax optimal rate.
\citet{2017arXiv170806633S} analyzed the same situation for the \Holder class which corresponds to $\beta^{(\ell)}_1=\dots=\beta^{(\ell)}_d~(\forall \ell)$ and $p = q = \infty$.
Our analysis far extends their analysis to the setting of anisotropic Besov spaces in which the parameters $\beta^{(\ell)},p,q$ have much more freedom.

From these two bounds (Theorems \ref{thm:EstimationErrorNNBesov_affine} and \ref{thm:EstimationErrorNNBesov_Deep}), we can see that as the smoothness $\betatild$ becomes large, the convergence rates faster.
If the target function is included in the isotropic Besov space with smoothness $\beta_1 = \dots = \beta_d (= \betamin)$,
then the estimation error becomes
\begin{flalign*}
~~~~\text{\bf (Isotropic Besov)} & &\textstyle n^{-2 \betamin/(2\betamin + d)}. %n^{-\frac{2 \betamin}{2\betamin + d}}.
 ~~~~~~~~&  ~~~~~~~~&
\end{flalign*}
In the exponent, the dimensionality $d$ appears, which causes the curse of dimensionality.
In contrast, if the target function is in the anisotropic Besov space, and the smoothness in each direction is sufficiently imbalanced such that $\betatild$ does not depend on $d$,
our obtained rate 
%$$n^{-\frac{2 \betatild}{2\betatild + 1}}$$ 
\begin{flalign*}
~~~~\text{\bf (Anisotropic Besov)} & &\textstyle n^{-2 \betatild/(2\betatild + 1)}  %n^{-\frac{2 \betatild}{2\betatild + 1}} 
~~~~~~~~~~~~~~~&  ~~~~~~~~&
\end{flalign*}
avoids the curse of dimensionality.
For high-dimensional settings, there would be several redundant directions in which the true function does not change.
Deep learning is adaptive to this redundancy and achieves a better estimation error.
%This explains why deep learning can generalize in practice.
However, in Section \ref{sec:MinimaxRateOfLinearEstimator}, we prove that linear estimators are affected by the dimensionality more strongly than deep learning.
This indicates the superiority of deep learning.
%Another important point is that deep learning achieves the {\it adaptive estimation error} rate.
%%For simplicity let $d=1$, then 
%It will be shown that the linear estimators can achieve only the rate 
%$
%n^{- \frac{2\betatild - 2(1/p-1/2)_+}{2 \betatild +1 - 2(1/p-1/2)_+}}.
%$
%Since there is an additional term $2(1/p-1/2)_+$ in the exponent, this results in a sub-optimal rate, whereas deep learning achieves the optimal rate.
%This is due to the {\it adaptivity} of deep learning, which was reported by \citet{suzuki2018adaptivity}
%(see discussions after Theorem \ref{thm:MinimaxLinear} for more details).

%in Section \ref{sec:ApproximatoinErroBounds} for more details).
%;
%that is, deep learning corresponds to a method that constructs basis functions depending on the target function
%while the linear estimators such a kernel method does not change the basis function depending on the target function

\vspace{-0.1cm}
\section{Minimax optimal rate}
\label{sec:MinimaxOpt}
\vspace{-0.1cm}

Here, we show that the estimation error rate, that we have presented, of deep learning achieves the {\it minimax optimal rate}.
Roughly speaking the minimax optimal risk on a model $\calF^\circ$ of the true function is the smallest worst case error over all estimators:
\begin{align*}
\textstyle
R_{*}(\calF^\circ) := \inf_{\fhat} \sup_{\ftrue \in \calF^\circ} \EE_{D_n}[\|\fhat - \ftrue \|_{\LPiPx}^2],
\end{align*}
where $\fhat$ runs over all estimators.
The convergence rate of the minimax optimal risk is referred to as minimax optimal rate.
We obtain the following minimax optimal rate for anisotropic Besov spaces.

\begin{Theorem}
\label{thm:MinimaxRate}
{\bf (a) Affine composition model:} For $0 < p,q \leq \infty$ and $\beta \in \Real_{++}^d$, assume that 
$
\betatild > \max\left\{1/p - 1/2,1/p -1 ,0\right\}.
$
Then, the minimax optimal risk of the affine composition model is lower bounded as 
$
R_*(\calHaff) \gtrsim n^{- \frac{2\betatild}{2 \betatild + 1}}.
$
{\bf (b) Deep composition model:} For $0 < p,q \leq \infty$ and $\beta^{(\ell)} \in \Real_{++}^d~(\ell=1,\dots,H)$, assume that 
$
\betatild^{(\ell)} > 1/p.
$
Let $\epsilon > 0$ be arbitrarily small for $q < \infty$, and let $\epsilon = 0$ for $q=0$.
Let $\betatild^{*(\ell)} = \betatild^{(\ell)} \prod_{k=\ell+1}^H [(\betamin^{(k)} -1/p + \epsilon)\wedge 1]$,
and $\betatild^{**} := \min_\ell \betatild^{*(\ell)}$.
Then, the minimax optimal risk of the deep composition model is lower bounded as 
$
R_*(\calHdeep) \gtrsim n^{- \frac{2\betatild^{**}}{2 \betatild^{**} + 1}}.
$
\end{Theorem}
The proof  is provided in Appendix \ref{sec:MinimaxMixedSmooth}
(see also \citet{ibragimov1984more,nyssbaum1987nonparametric}).
From this theorem, we can see that the estimation error of deep learning shown in Theorems \ref{thm:EstimationErrorNNBesov_affine} and \ref{thm:EstimationErrorNNBesov_Deep} indeed achieve the minimax optimal rate up to a poly-$\log(n)$ factor. % with arbitrarily small modification of $\betatild^{*(\ell)}$.

\section{Suboptimality of %kernel ridge regression and 
linear estimators}
\label{sec:MinimaxRateOfLinearEstimator}

In this section, we give the minimax optimal rate in the class of {\it linear estimators}.
The linear estimator is a class of estimators that can be written as
\begin{align*}
\textstyle
\fhat(x) = %\varphi(x;X^n) + 
\sum_{i=1}^n y_i \varphi_i(x;X^n),
\end{align*}
where $X^n = (x_1,\dots,x_n)$ and $\varphi_i(x;X^n)~(i=1,\dots,n)$ are (measurable) functions that only depend on $x$ and $X^n$.
%\footnote{This estimator probably should be called ``affine estimators'' but we employ the terminology ``linear estimator'' following convention.}
This is linearly dependent on $Y^n=(y_1,\dots,y_n)$.
We notice that the kernel ridge regression is included in this class because it can be written as
$\fhat(x) = k_{x,X^n}(k_{X^n,X^n} + \lambda \mathrm{I})^{-1}Y^n$, which linearly depends on $Y^n$.
This class includes other important estimators, such as the Nadaraya--Watson estimator, the $k$-nearest neighbor estimator, and the sieve estimator.
We compare deep learning with the linear estimators in terms of minimax risk. 
For this purpose, we define the minimax risk of the class of linear estimators:
$$
%\textstyle
R_*^{(\mathrm{lin})}(\calF^\circ) := 
\inf_{\text{$\fhat$: linear}} \sup_{\ftrue \in \calF^\circ}\EE_{D_n}[\|\ftrue -  \fhat\|_{\LPi(P_X)}^2],
$$
where $\fhat$ runs over all {\it linear estimators}.
We can see that linear estimators suffer from the sub-optimal rate because of the following two points: 
(i) they do not have adaptivity, and (ii) they significantly suffer from the curse of dimensionality.

\begin{Theorem}
\label{thm:MinimaxLinear}
%Let us consider the affine composition model. 
(i) Suppose that the input distribution $P_X$ is uniform distribution on $\Omega = [0,1]^d$ and $\betatild > 1/p$, $1 \leq p,q \leq \infty$.
Then, the minimax optimal rate of the linear estimators is lower bounded as 
\begin{align}
R_*^{(\mathrm{lin})}(U(B_{p,q}^\beta))
\gtrsim n^{- \frac{2 \betatild - v }{2\betatild + 1 - v}},
\label{eq:MinimaxLinearNonAdaptive}
\end{align}
where $v = 2(1/p - 1/2)_+$.
(ii) In addition to the above conditions, we assume that $\dtil \leq d$, % and 
%there exists $\tilde{d}$ such that 
$\betamin = \beta_1 = \dots = \beta_{\dtil}$ and $0 < p \leq 2$. 
Let $a_d = 1 +\kappa$ (with arbitrary small $\kappa > 0$) %\footnote{\scriptsize Correction from the main submission.} 
when $\dtil<d/2$ and $a_d = 0$ when $\dtil \geq d/2$.
Then, the minimax rate of the linear estimators on the affine composition model is lower bounded by
\begin{align}
R_*^{(\mathrm{lin})}(\calHaff) 
\gtrsim n^{- \frac{2 (\betamin - \dtil/p  + d/2 + a_d )}{2 (\betamin - \dtil/p + d/2+ a_d ) + d}}.
%n^{- \frac{2 (\betamin - \dtil/p + d/2)}{2 (\betamin - \dtil/p  + d/2) + d}}.
\label{eq:MinimaxLinearAffineHull}
\end{align}
\end{Theorem}
The proof is provided in Appendix \ref{sec:MinimaxLinearProof}.
(i) The lower bound \eqref{eq:MinimaxLinearAffineHull} reveals the suboptimality of linear estimators in terms of input dimensionality. 
Actually, if we consider a particular case where $\dtil = 1$, $p=1$ and $d \gg \dtil$, %assume that $d$ is even, $\dtil = d/2$ and $p=1$, 
then the obtained minimax rate of linear estimators and the estimation error rate of deep learning can be summarized as  %is simplified as 
$$
%\text{linear}:~n^{- \frac{2 \betamin}{2 \betamin + 2d}},\qquad\quad\text{deep}:~n^{- \frac{2 \betamin}{2 \betamin + d/2}},
\text{linear}:~n^{- \frac{2 \betamin + d}{2 \betamin + 2d}},\qquad\quad\text{deep}:~n^{- \frac{2 \betamin}{2 \betamin +1}},
$$
%whereas the estimation error rate of deep learning is bounded by
%$$
%n^{- \frac{2 \betamin}{2 \betamin + d/2}}
%$$
by Theorem \ref{thm:EstimationErrorNNBesov_affine} when $\betamin > 1$ (which can be checked by noticing $\dtil = \betamin/\betatild=1$ in this situation).
We can see that the dependence on the dimensionality of linear estimators is significantly worse than that of deep leaning. 
This indicates poor adaptivity of linear estimators to the intrinsic dimensionality of data. 
Actually, as $d$ becomes large, the rate for the linear estimator approaches to $1/\sqrt{n}$ but that for the deep learning is not affected by $d$ and still faster than $1/\sqrt{n}$.
To show the theorem, we used the ``convex-hull argument'' 
developed by \citet{HayakawaSuzuki2019deep,donoho1998minimax}.
We combined this technique with the so-called {\it Irie-Miyake's integral representation} \citep{ICNN:IrieMiyake:1988,NN:HORNIK:1990551}.
%; that is, the minimax lower bound of linear estimators is same as that on the convex hull of the target function class. 
%Based on this observation, linear estimators must behave as if the target function class is a convex hull of the affine composition model. 
%The property that the target function is dependent only on particular few directions of the input space is not preserved if we take the convex hull, resulting in the suboptimality of linear estimators. 
Note that this difference appears because there is an affine transformation in the first layer of the affine composition model.
Deep learning is flexible against such a coordinate transform so that it can find directions to which the target function is smooth.
In contrast, kernel methods do not have such adaptivity because there is no feature extraction layer.
%If we allow $p$ to be dependent on the input dimensionality $d$, 
%then for a situation where $p = 2/d$ and $\dtil = 1$, 
%the minimax rate of the linear estimator is lower bounded by
%$
%n^{- \frac{2 s}{2 s + d}},
%$
%but deep learning achieves the estimation error bound $n^{- \frac{2 s}{2 s + 1}}$ where $s = \beta_1 (=\betatild = \betamin=\betamax) \in \Real_{++}$ for $\dtil = 1$.
%In this situation, the effect of $d$ is clearly highlighted.
%However, we should notice that to satisfy $\betatild > (1/p - 1/2)_+$,~$\betatild$ should also be dependent on $d$ as $\betatild > (d-1)/2$ for $p=2/d$.
%Therefore, the effect of $d$ to the convergence rate is not as strong as the situation with fixed $p$.
(ii) The lower bound \eqref{eq:MinimaxLinearNonAdaptive} states that when $p < 2$ (that is, $v > 0$), 
the minimax rate of the linear estimators is outperformed by that of deep learning (Theorem \ref{thm:EstimationErrorNNBesov_affine}).
This is due to the ``adaptivity'' of deep leaning.
When $p$ is small, the smoothness of the target function is less homogeneous, and it requires an adaptive approximation scheme to achieve the best estimation error.  Linear estimators do not have adaptivity and thus fail to achieve the minimax optimal rate. % while deep learning can. %This is reported by \citet{suzuki2018adaptivity} for the isotropic Besov space. 
%This argument can go back to \cite{donoho1998minimax} who discussed the suboptimality of linear estimators comparing with wavelet shrinkage estimators.
Our bound \eqref{eq:MinimaxLinearNonAdaptive} extends the result by \citet{zhang2002wavelet} to a multivariate anisotropic Besov space while \citet{zhang2002wavelet} investigated the univariate space ($d=1$).

%\begin{Remark}
%\citet{2017arXiv170806633S} considered a set of functions of the form of $
%\ftrue(x) = g(w^\top x)
%$
%where $g \in \calC^{\beta}([0,1])$ and $w \in \Real^d$,
%and showed the sub-optimailty of a wavelet estimator, which is one of linear estimators.
%It is shown that the minimax optimal rate of a certain class of wavelet estimators is lower bounded by
%$n^{-2 \betamin/(2 \betamin + d)}$. 
%This is a stronger result than our bound \eqref{eq:MinimaxLinearNonAdaptive}.
%This is because the above rate is proven for a specific wavelet estimator, whereas ours can be applied to {\it ``any''} linear estimator.
%%Actually, they investigate a specific class of functions to derive the lower bound for which we can show that there exists a linear estimator that achieves the optimal rate $n^{-\frac{2 \betamin}{2 \betamin + 1}}$.
%%However, we also conjecture that our bound \eqref{eq:MinimaxLinearNonAdaptive} could be further improved.  We leave this issue for future work.
%%We would like to leave this issue for future work.
%\end{Remark}
\section{Conclusion}\label{sec:Conclusion}

We investigated the approximation error and estimation error of deep learning in the anisotropic Besov spaces. 
It was proved that the convergence rate is determined by the average of the anisotropic smoothness, which results in milder dependence on the input dimensionality. If the smoothness is highly anisotropic, deep learning can avoid overfitting.
We also compared the error rate of deep learning with that of linear estimators and showed that deep learning has better dependence on the input dimensionality.
Moreover, it was shown that deep learning can achieve the adaptive rate and outperform non-adaptive approximation methods and linear estimators if the homogeneity $p$ of smoothness is small. 
These analyses strongly support the practical success of deep learning from a theoretical perspective.

\vspace{-0.2cm}
\paragraph{Limitations of this work}
Our work does not cover the optimization aspect of deep learning. 
It is assumed that the regularized least squares \eqref{eq:ERMprocedure} can be executed. 
It would be nice to combine our study with recent developments of non-convex optimization techniques \citep{vempala2019rapid,suzuki2021benefit}.

\vspace{-0.2cm}
\paragraph{Potential negative societal impact}
Since this is purely theoretical result, it is not expected that there is a direct negative societal impact.
However, revealing detailed properties of the deep learning could promote an opportunity to pervert deep learning.

\section*{Acknowledgment}
TS was partially supported by JSPS KAKENHI (18H03201), Japan Digital Design and JST CREST.
AN was partially supported by JSPS Kakenhi (19K20337) and JST-PRESTO.

\bibliographystyle{iclr2019_conference_mod.bst}
\bibliography{main,main_colt}

\appendix

\begin{comment}
\section{Another approximation error bound for cardinal B-splines}

\begin{Proposition}[\cite{chui1994neural}]
\label{prop:ChuiMhaskarMapprox}
Assume that the activation function satisfies \Eqref{eq:activation_kterm} with $k \geq 2$.
%\footnote{Note that this situation excludes  %ReLU activation.}.
Let 
$L = \lceil \log(md -d )/\log(k)\rceil + 1$,
and $W = 2 d m^d (k+1) {k^L +d  \choose d}$. 
Then, for all $\epsilon > 0$, 
there exists $S_\epsilon$ depending on $\epsilon$
such that there exists a neural network 
$
\check{M} \in \Phi(L,W,S_\epsilon,B) 
$ satisfying 
$$
\|M_{0,0}^d - \check{M}\|_{L^\infty([0,1]^d)} \leq \epsilon.
$$
%In particular, 
\end{Proposition}
\end{comment}

\onecolumn
\appendix

\begin{center}
{\bf \LARGE ------Appendix------}
\end{center}

\section{Approximation error of Affine composition model}\label{sec:AffineCompApproxError}
% %\paragraph{Affine composition model}
\begin{Theorem}[Affine composition model]
\label{thm:AffineApproximationError}
Assume that the distribution of $\tilde{x} = A x +b \in \Real^{\dtil}$ has a bounded density function on $[0,1]^{\dtil}$ when $x$ obeys the uniform distribution on $\Omega$, and
each element in $A$ and $b$ is bounded by a constant $C$. Assume that
$0 < p,q,r \leq \infty$ and $\beta \in \Real_{++}^{\dtil}$ satisfy 
$
\betatild > (1/p - 1/r)_{+}.
$
Then, it holds that
\begin{align}\label{eq:AffineCompApprox}
& %\textstyle 
R_r(\Phi(L_1(\dtil),W_1(\dtil),S_1(\dtil),(\dtil C +1)B_1(\dtil)),\calH_{\mathrm{aff}}) \lesssim N^{-\betatild},
%\\ & 
\end{align}
where $L_1(\cdot),~W_1(\cdot),~S_1(\cdot)$,~$B_1(\cdot)$ are defined in \Eqref{eq:L1B1def}.
\end{Theorem}
The assumption $\betatild > (1/p - 1/r)_{+}$ ensures the $L_r$-integrability of the target function, and the inequality (without equality) admits a near-optimal wavelet approximation of the target function in terms of $L_r$-norm.
From this theorem, the approximation error is almost identical to that for $B^\beta_{p,q}(\Omega)$ (Proposition \ref{prop:BesovApproxByNN}).

\section{Proofs of approximation error bounds}
\label{sec:ProofPropBesovApp}

To show the approximation accuracy, 
a key step is to show that the ReLU neural network can approximate the {\it cardinal B-spline} with high accuracy.
Let $\calN(x) =  1~  (x \in [0,1]),~0 ~ (\text{otherwise})$, then the 
{\it cardinal B-spline of order $m$} is defined by taking $m+1$-times convolution of $\calN$:
$$
\calN_m(x) = (\underbrace{\calN * \calN * \dots * \calN}_{\text{$m+1$ times}})(x),
$$
where $f* g(x) := \int f(x -t) g(t) \dd t$. It is known that $\calN_m$ is a piece-wise polynomial of order $m$.
For $k \in \Integer_+^d$ and $j=(j_1,\dots,j_d)\in \Integer_+^d$, let $$
M_{k,j}^d(x) = \prod_{i=1}^d \calN_m(2^{\lfloor k \beta'_i\rfloor} x_i - j_i),
$$
where $\beta \in \Real_{++}^{d}$ is a given smoothness parameter (we omit the dependency on $\beta$ from the notation which would be obvious from the context).
%Even for $k \in \Integer_+$, we also use the same notation to express $M_{k,j}^d(x) = \prod_{i=1}^d \calN_m(2^{k} x_i - j_i)$.
Here, $k$ controls the spatial ``resolution'' and $j$ specifies the location on which the basis is put.
Basically, we approximate a function $f$ in an anisotropic Besov space via super-position of $M_{k,j}^m(x)$, which is closely related to wavelet analysis \citep{Mallat99a}. The following is a key lemma that was proven by \citet{suzuki2018adaptivity}.

\begin{Lemma}[Approximation of cardinal B-spline basis by the ReLU activation]
\label{lemm:Mnapproximation}
%Let 
%$L := \lceil \log(md -d )/\log(k)\rceil + 1$,
%and $W := 2 d m^d (k+1) {k^L +d  \choose d}$. 
There exists a constant $c_{(d,m)}$ depending only on $d$ and $m$
such that,
for all $\epsilon > 0$,
%there exists $S_\epsilon$ depending on $\epsilon$ such that 
there exists a neural network 
$
\check{M} \in \Phi(L_0,W_0,S_0,B_0) 
$ with 
$L_0 := 3 +  2 \left\lceil \log_2\left(  \frac{3^{d\vee m} }{\epsilon c_{(d,m)}}  \right)+5 \right\rceil \left \lceil\log_2(d \vee m)\right\rceil$, 
$W_0 := 6 d m(m+2) + 2 d$, $S_0 := L_0 W_0^2$ and  $B_0 := 2 (m+1)^m$
that 
satisfies
$$
\|M_{0,0}^d - \check{M}\|_{L^\infty(\Real^d)} \leq \epsilon,
$$
and $\check{M}(x) = 0$ for all $x \not \in [0,m+1]^d$.
%In particular, 

\end{Lemma}

Let 
$$
\|k\|_{\betamin/\beta} := \sum_{j=1}^d \lfloor k \betamin/\beta_j \rfloor
$$
for a $k \in \Integer$.
For order $m \in \Natural$ of the cardinal B-spline bases, let 
$$J_i(k) = \{- m,  - m +1, \dots, 2^{\lfloor k \beta'_i \rfloor}-1, 2^{\lfloor k \beta'_i \rfloor} \}$$
and 
$$
J(k):= J_1(k) \times J_2 (k) \times \dots \times J_d(k).
$$ 
and 
the quasi-norm of the coefficient $(\alpha_{k,j})_{k,j}$ for $k \in \Integer_+$ and $j \in J(k)$ be 
$$
\| (\alpha_{k,j})_{k,j}\|_{b^\beta_{p,q}}. =  \left\{\sum_{k =0}^\infty \left[ 2^{k[\betamin - (\sum_{i=1}^d \lfloor k\beta'_i\rfloor/k) /p]}\Big(\sum_{j \in J(k)} |\alpha_{k,j}|^p\Big)^{1/p}  \right]^q\right\}^{1/q}.
$$
For $p=\infty$ or $q=\infty$, the definition should be appropriately modified as usual.

\begin{Lemma}
\label{lemm:BSplineInterpolation}
Assume the condition $\betatild > (1/p - 1/r)_+$ %\eqref{eq:sprConditions} 
in Proposition \ref{prop:BesovApproxByNN}
and $0 < \betamax < \min( m , m -1 + 1/p)$ where $m \in \Natural$ is the order of the cardinal B-spline bases.
Then, $f \in B^\beta_{p,q}$ admits the following decomposition:
\begin{align}
\label{eq:DevorePopovExpansion}
f = \sum_{k=0}^\infty \sum_{j \in J(k)}  \alpha_{k,j} M_{k,j}^d(x)
\end{align}
with convergence in the sense of $L^p$, and the coefficient $(\alpha_{k,j})$ yields the following norm equivalence
%Moreover, we have that 
\begin{align}
\label{eq:BpqEquivalence}
\|f\|_{B_{p,q}^\beta} \simeq \| (\alpha_{k,j})_{k,j}\|_{b^\beta_{p,q}}.
\end{align}

For an integer $K \in \Natural$, let $N =  \lceil 2^{\|K\|_{\betamin/\beta}} \rceil$, 
then for any $f \in B^\beta_{p,q}(\Omega)$, there exists $f_N$ 
that satisfies
\begin{align}
%\label{eq:ffNoptimalAdaptiveApprox}
\notag
\|f - f_N\|_{L^r(\Omega)} \lesssim N^{-\betatild} \|f\|_{B^\beta_{p,q}},
\end{align}
%for $N \gg 1$, 
and has the following form:
\begin{align}
\label{eq:fNformat}
f_N(x) = \sum_{k=0}^K \sum_{j \in J(k)} \alpha_{k,j} M_{k,j}^d(x) + \sum_{k=K+1}^{K^*} \sum_{i=1}^{n_k}
\alpha_{k,j_i} M_{k,j_i}^d(x),
\end{align}
where 
%we define 
%$K = \lceil C_1 \log(N) (\sum_{j=1}^d \beta_j/\betamin)^{-1} \rceil$, 
$K^* = \lceil K(1 + 1/\nu) \rceil$,  
$n_k = \lceil 2^{ \|K\|_{\betamin/\beta} -\epsilon ( \|k\|_{\betamin/\beta} - \|K\|_{\betamin/\beta})}\rceil$ $(k = K+1, \dots,  K^*)$
for $\delta = (1/p - 1/r)_+$ and $\nu = %\epsilon = 
(\betatild - \delta)/(2\delta)$, %(s - \delta)/(2\delta)$ 
and 
$(j_i)_{i=1}^{n_k}\subset J(k)$.
%$$
%%C_1 2^{K} \leq N \leq C_2 2^K, 
%N =  \lceil 2^{\|K\|_{\betamin/\beta}} \rceil. %\lceil \log_2(\|K\|_{\betamin/\beta}) \rceil.
%$$
%%where $C_1$ and $C_2$ will be chosen properly later.
%Let $\epsilon = (\betatild - \delta)/(2\delta)$, and 
%$$
%K^* = \lceil K(1 + 1/\epsilon) \rceil, %\log(C_3 N)/\epsilon\rceil + K + 1,
%$$
%and $$n_k = \lceil 2^{ \|K\|_{\betamin/\beta} -\epsilon ( \|k\|_{\betamin/\beta} - \|K\|_{\betamin/\beta})}\rceil$$ for $k \in \Integer_+$ with $K+1 \leq k \leq  K^*$.
%$\delta = (1/p - 1/r)_+$ and $\epsilon = %\epsilon = 
%(\betatild - \delta)/(2\delta)$, %(s - \delta)/(2\delta)$ 
%we define 
%$K = \lceil C_1 \log(N) (\sum_{j=1}^d \beta_j/\betamin)^{-1} \rceil$, 
%$K^* = \lceil K(1 + 1/\epsilon) \rceil$, and 
%$n_k = \lceil 2^{ \|K\|_{\betamin/\beta} -\epsilon ( \|k\|_{\betamin/\beta} - \|K\|_{\betamin/\beta})}\rceil$ for $k = K+1, \dots,  K^*$.
%$K = \lceil C_1 \log(N)/d \rceil$, 
%$K^* = \lceil \log(\lambda N) \nu^{-1} \rceil + K + 1$,
%$n_k = \lceil \lambda N 2^{-\nu (k - K)}\rceil~
%(k=K+1,\dots,K^*)$
%for $\delta = (1/p - 1/r)_+$ and $\nu = 
%%\epsilon = 
%(\betatild - \delta)/(2\delta)$ %(s - \delta)/(2\delta)$ 
%where the real number constants $C_1 > 0$ and $\lambda > 0$ 
%are chosen to satisfy
%$\sum_{k=1}^K (2^k+m )^d + \sum_{k=K+1}^{K^*} n_k \leq N$ independently to $N$.

\end{Lemma}

\begin{proof}[Proof of Lemma \ref{lemm:BSplineInterpolation}]

%Let $J(k) := $

\cite{leisner2003nonlinear} %\cite{devore1988interpolation} % constructed an decomposition method of $f \in B_{p,q}^s$ based on a spline interpolant.
showed that there exists a bounded linear operator $P_k$ that %having the following form:
can be expressed as 
\begin{align}
P_k(f)(x) = \sum_{j \in J(k)} a_{k,j} M_{k,j}^d(x)
\label{eq:PkDecomposition}
\end{align}
where $\alpha_{k,j}$ is constructed in a certain way, and 
for every $f \in L^p([0,1]^d)$ with $0 < p \leq \infty$, it holds 
\begin{align*}
\| f - P_k(f) \|_{L^p} \leq C w_{r,p}(f,(2^{-k \beta'_1},\dots,2^{-k \beta'_d})),
%\label{eq:fPkwkpIneq}
\end{align*}
(See Theorem 3.2.4 of \citet{leisner2003nonlinear} and \citet{devore1988interpolation}).
Let
$$
p_k(f) := P_k(f) - P_{k-1}(f),~~P_{-1}(f) = 0.
$$
Then, \cite{leisner2003nonlinear} showed that when $0 < p, q \leq \infty$ and $0 < \betamax < \min( m , m -1 + 1/p)$, 
$f$ belongs to $B_{p,q}^\beta$ if and only if $f$ can be decomposed into 
$$
f = \sum_{k=0}^\infty p_k(f),
$$
with the convergence condition 
$$
\|(p_k(f))_{k=0}^{\infty} \|_{b_q^\beta(L^p)} := \left[\sum_{k =0}^\infty (2^{\betamin k} \|p_k\|_{L^p} )^q\right]^{1/q} < \infty.
$$
In particular, it is shown that 
\begin{align}
\label{eq:BpqNormEquivPk}
\|f\|_{B_{p,q}^s} \simeq \| (p_k(f))_{k=0}^{\infty} \|_{b_p^s(L^p)}.
\end{align}
Here, each $p_k$ can be expressed as $p_k(x) = \sum_{j \in J(k)} \alpha_{k,j} M_{k,j}^d(x)$ 
for a coefficient $(\alpha_{k,j})_{k,j}$ which could be different from  $(a_{k,j})_{k,j}$ appearing in \Eqref{eq:PkDecomposition},
and thus $f \in B^\beta_{p,q}$ can be decomposed into 
\begin{align*}
%\label{eq:DevorePopovExpansion}
f = \sum_{k=0}^\infty \sum_{j \in J(k)}  \alpha_{k,j} M_{k,j}^d(x)
\end{align*}
with convergence in the sense of $L^p$. %which yields \Eqref{eq:DevorePopovExpansion}.
Moreover, it is shown that $\|p_k\|_{L^p} \simeq( 2^{-kd}\sum_{j \in J(k)} |\alpha_{k,j}|^p  )^{1/p}$ and thus
\begin{align*}
\|f\|_{B_{p,q}^\beta} \simeq \| (\alpha_{k,j})_{k,j}\|_{b^\beta_{p,q}}.
\end{align*}
This yields the first assertion.

Next, we move to the second assertion.
If $p \geq r$, the assertion can be shown in the same manner 
as Theorem 3.1 of \cite{Complexity:Dung:2011}.
More precisely, we can show the assertion in a similar line to the following proof for $p < r$ by setting $K = K^*$.
Thus, we show the assertion only for $p < r$. 
In this regime, we need to use an adaptive approximation method.
In the following, we assume $p < r$.
For a given $K$, by appropriately choosing $K^*$ later, 
we set 
$$
R_K(f)(x) = \sum_{0 \leq k \leq K}%k \in \Integer_+^d: \|k\|_1 \leq K} 
p_k
+ \sum_{k \in \Integer_+: K < k \leq K^*} G_k(p_k),
$$
where $G_k(p_k)$  is given as 
$$
G_k(p_k)  = \sum_{1 \leq i \leq n_k} \alpha_{k,j_i} M_{k,j_i}^d(x)
$$
where $(\alpha_{k,j_i})_{i=1}^{|J(k)|}$
is the sorted coefficients in decreasing order of absolute value: 
$|\alpha_{k,j_1}| \geq |\alpha_{k,j_2}|
\geq \dots \geq |\alpha_{k,j_{|J(k)|}}|$.
%for $p_k = \sum_{j \in J_m^d(k) } \alpha_{k,j} M_{k,j}^d(x)$
Then, it holds that 
$$
\|p_k - G_k(p_k)\|_{r} \leq \|p_k\|_{p} 2^{\delta \|k\|_{\betamin/\beta}} n_k^{-\delta},
$$
where $\delta := (1/p - 1/r)$ (see the proof of Theorem 3.1 of \cite{dung2011optimal}
and Lemma 5.3 of \cite{Complexity:Dung:2011}).
Moreover, we also have 
$$
\|p_k \|_{r} \leq \|p_k\|_{p} 2^{\delta \|k\|_{\betamin/\beta}} 
$$
for $k \in \Integer_+$ with $k > K^*$.

Here, we define $N$ as 
$$
%C_1 2^{K} \leq N \leq C_2 2^K, 
N =  \lceil 2^{\|K\|_{\betamin/\beta}} \rceil. %\lceil \log_2(\|K\|_{\betamin/\beta}) \rceil.
$$
%where $C_1$ and $C_2$ will be chosen properly later.
Let $\nu = (\betatild - \delta)/(2\delta)$,  
$$
K^* = \lceil K(1 + 1/\nu) \rceil, %\log(C_3 N)/\epsilon\rceil + K + 1,
$$
and 
$$
n_k = \left\lceil 2^{ \|K\|_{\betamin/\beta} -\epsilon ( \|k\|_{\betamin/\beta} - \|K\|_{\betamin/\beta})} \right\rceil
$$ 
for $k \in \Integer_+$ with $K+1 \leq k \leq  K^*$.
%The constants $0 < C_1,C_2,C_3$ are chosen to satisfy
%$$
%2^K + \sum_{k: K < \|k\|_1 \leq K^*} n_k \leq N !!!!!!!!!!!!!!!
%$$

Then, by Lemma 5.3 of \cite{Complexity:Dung:2011}, we have 
\begin{align}
\|f - R_K(f)\|_{L^r}^r 
& \lesssim \sum_{K  < k \leq K^*} 
%[2^{\delta \|k\|_1} \|p_k - G_k(p_k)\|_{L^p}]^r
 \|p_k - G_k(p_k)\|_{L^r}^r
+ 
\sum_{K^*  < k }  \|p_k \|_{L^r}^r
\notag
\\
& 
\lesssim \sum_{K  < k \leq K^*} 
%[2^{\delta \|k\|_1} \|p_k - G_k(p_k)\|_{L^p}]^r
[ \|p_k\|_{p} 2^{\delta \|k\|_{\betamin/\beta}} n_k^{- \delta}]^r
+ 
\sum_{K^*  < k } 
[2^{\delta \|k\|_{\betamin/\beta}} \|p_k \|_{L^p}]^r.
\label{eq:fRNsubtraction}
\end{align}

%In the following, we require an upper bound of 
%${k + d-1 \choose d-1 }$.
%Hence, we evaluate this quantity beforehand. 
%This can be upper bounded 
%by the Stering's formula as 
%$${k + d-1 \choose d-1 } \leq \frac{\sqrt{2}e}{2 \pi} 
%\underbrace{\left(1 + \frac{d-1}{k}\right)^k \left(1 + \frac{k}{d-1}\right)^{d-1}}_{= D_{k,d}} \leq D_{k,d}.$$
%Let $\xi > 0$ be a positive real number satisfying  $1 + \xi \geq K^*/K$. %$\xi$ can be chosen independently of $N$.
%We can see that $\xi$ can be chosen as $\xi =  1/\epsilon + o(1)$.
%Then, we have that 
%\begin{align}
%D_{K^*,d} & = D_{K,d}
%\frac{(1 + \frac{d-1}{K^*})^{K^*}}{(1 + \frac{d-1}{K})^{K}}
%\frac{(1 + \frac{K^*}{d-1})^{d-1}}{(1 + \frac{K}{d-1})^{d-1}}
%\leq 
%D_{K,d} \frac{(1 + \frac{d-1}{K})^{K^*}}{(1 + \frac{d-1}{K})^{K}}
%\left(\frac{1}{1 + \frac{K}{d-1}} + \frac{K^*}{(d-1)
%(1 + \frac{K}{d-1})}\right)^{d-1} \notag \\
%&
%\leq
%D_{K,d} \left(1 + \frac{d-1}{K}\right)^{K^*-K}
%\left(\frac{d -1 + K^*}{d-1 + K}\right)^{d-1} 
%=
%D_{K,d} 
%\left(1 + \frac{d-1}{K}\right)^{\xi K}
%\left(1 + \xi\right)^{d-1} \notag \\
%& \leq 
%D_{K,d} 
%e^{(d-1)\xi} 
%(1 + \xi)^{d-1} \simeq D_{K,d}.
%\label{eq:Dkd_upperbound}
%\end{align}

(a) Suppose that $q \leq r$ and $r<\infty$. Then,
\begin{align*}
& \|f - R_K(f)\|_{L^r}^q =\|f - R_K(f)\|_{L^r}^{r \frac{q }{r}}  \\
& \lesssim \left\{ \sum_{K  < \|k\|_1 \leq K^*} 
%[2^{\delta \|k\|_1} \|p_k - G_k(p_k)\|_{L^p}]^r
[2^{\delta \|k\|_{\betamin/\beta} } n_k^{-\delta} \|p_k \|_{L^p}]^r
+ 
\sum_{K^*  < k} 
[2^{\delta \|k\|_{\betamin/\beta}} \|p_k \|_{L^p}]^r \right\}^{\frac{q}{r}} ~~~~~~~~(\text{$\because$ \Eqref{eq:fRNsubtraction}}) \\
& \lesssim \sum_{K  < k \leq K^*} 
%[2^{\delta \|k\|_1} \|p_k - G_k(p_k)\|_{L^p}]^r
[2^{\delta \|k\|_{\betamin/\beta} } n_k^{-\delta} \|p_k \|_{L^p}]^q
+ 
\sum_{K^*  < k } 
[2^{\delta \|k\|_{\betamin/\beta}} \|p_k \|_{L^p}]^q \\
& \leq
N^{- \delta q} 2^{-(\betatild- \delta) \|K\|_{\betamin/\beta} q}
\sum_{K  < k \leq K^*} 
%[2^{\delta \|k\|_1} \|p_k - G_k(p_k)\|_{L^p}]^r
[\underbrace{2^{-(\betatild - \delta - \delta \epsilon)( \|k\|_{\betamin/\beta} - \|K\|_{\betamin/\beta})}}_{\leq 1} 
2^{\betatild \|k\|_{\betamin/\beta}} \|p_k \|_{L^p}]^q \\
& ~~~+
2^{-q (\betatild - \delta) \|K^*\|_{\betamin/\beta}} 
\sum_{K^*  < k } [2^{\betatild \|k\|_{\betamin/\beta}} \|p_k \|_{L^p}]^q \\
& \overset{\text{(i)}}{\lesssim}
(N^{- \delta} 2^{-(\betatild - \delta)\|K\|_{\betamin/\beta}} 
+2^{- (\betatild - \delta) K^*} )^q
\|f\|_{\MB^s_{p,q}}^q 
~~~~(\text{$\because$ \Eqref{eq:BpqNormEquivPk}})
\\
& \overset{\text{(ii)}}{\lesssim} 
(N^{- \betatild} )^q
\|f\|_{\MB^\alpha_{p,q}}^q
\end{align*}
where we used $2^{\betatild \|k\|_{\betamin/\beta}} \simeq 2^{\betamin k}$ in (i),  and 
$N \simeq 2^{ \|K\|_{\betamin/\beta}}$ and $\nu = (\betatild - \delta)/(2\delta)$ in (ii).
%Hence, we it holds that 
%
%$p_k = \sum_{j }\alpha_{k,j_i} M_{k,j_i}^d(x)$
%\begin{align}
%P_k(f)(x) = \sum_{j \in J(k)} a_{k,j} M_{k,j}^d(x)
%\ref{eq:PkDecomposition}
%\end{align}

(b) Suppose that $q > r$ and $r < \infty$. Then,
letting $\gamma = q/r ~( > 1)$ and $\gamma' = 1/(1-1/\gamma) = q/(q - r)$ (note that $\frac{1}{\gamma} + \frac{1}{\gamma'} = 1$), we have 
\begin{align*}
& \|f - R_K(f)\|_{L^r}^r
\lesssim \sum_{K  < k \leq K^*} 
%[2^{\delta \|k\|_1} \|p_k - G_k(p_k)\|_{L^p}]^r
[2^{\delta \|k\normb } n_k^{-\delta} \|p_k \|_{L^p}]^r
+ 
\sum_{K^*  < k } 
[2^{\delta \|k\normb} \|p_k \|_{L^p}]^r ~~~~~(\text{$\because$ \Eqref{eq:fRNsubtraction}})\\
%%%%%%%%%%%%%%%%%%%%%%%%%%
& \leq
 2^{-\betatild \|K\normb r}
\sum_{K  < k \leq K^*} 
%[2^{\delta \|k\|_1} \|p_k - G_k(p_k)\|_{L^p}]^r
[2^{-(\betatild - \delta - \delta \nu)( \|k\normb - \|K\normb)} 2^{\betatild \|k\normb} \|p_k \|_{L^p}]^r
\\& ~~~~~~~+
\sum_{K^*  < k} [2^{\betatild \|k\normb} \|p_k \|_{L^p}]^r
(2^{- (\betatild - \delta)\|k\normb} )^r
 \\
%%%%%%%%%%%%%%%%%%%%%%%%%%
& \leq
(2^{-\betatild\|K\normb} 
+2^{- (\betatild - \delta) \|K^*\normb} )^r
\Big\{
\sum_{K  < k \leq K^*} 
%[2^{\delta \|k\|_1} \|p_k - G_k(p_k)\|_{L^p}]^r
[2^{-(\betatild - \delta - \delta \nu)( \|k\normb - \|K\normb)} 2^{\betatild \|k\normb} \|p_k \|_{L^p}]^r  \\
& +
\sum_{K^*  < k } [2^{\betatild \|k\normb} \|p_k \|_{L^p}]^r
2^{-(\betatild -\delta)(\|k\normb- \|K^*\normb)r}
\Big\}
\\
& \leq 
(2^{-\betatild \|K\normb r}
+2^{- (\betatild - \delta) \|K^*\normb} )^r
\left\{
\sum_{K  < k \leq K^*} 
%[2^{\delta \|k\|_1} \|p_k - G_k(p_k)\|_{L^p}]^r
[ 2^{\betatild \|k\normb} \|p_k \|_{L^p}]^{r\gamma} 
+
\sum_{K^*  < k} [2^{\betatild \|k \normb} \|p_k \|_{L^p}]^{r\gamma}
\right\}^{1/\gamma}\\
&
\times
\left\{
\sum_{K  < k \leq K^*} 
%[2^{\delta \|k\|_1} \|p_k - G_k(p_k)\|_{L^p}]^r
[2^{-(\betatild - \delta - \delta \nu)( \|k\normb - \|K\normb) }]^{r \gamma'} 
%2^{\alpha \|k\|_1} \|p_k \|_{L^p}]^r 
+
\sum_{K^*  < k } [2^{-(s -\delta)(\|k\normb - K^*) }]^{r \gamma'}
\right\}^{1/\gamma'}
\\
& \lesssim
( 2^{-\betatild \|K\normb} 
+2^{- (\betatild - \delta) \|K^*\normb} )^r
\|f\|_{B^\beta_{p,q}}^r % D_{K,d}^{r(1/r - 1/q)}
%~~~~~~~~~(\because \text{\Eqref{eq:Dkd_upperbound}})
~~~~(\text{$\because$ \Eqref{eq:BpqNormEquivPk} and $2^{\betatild \|k\|_{\betamin/\beta}} \simeq 2^{\betamin k}$ })
\\
& \lesssim
(N^{- \betatild})^r \|f\|_{B^\beta_{p,q}}^r.
\end{align*}

(c) Suppose that $r=\infty$.
Then, similarly to the analysis in (b), we can evaluate 
\begin{align*}
& \|f - R_K(f)\|_{L^r} \\
& \lesssim 
 2^{-\betatild \|K\normb }
\sum_{K  < k \leq K^*} 
%[2^{\delta \|k\|_1} \|p_k - G_k(p_k)\|_{L^p}]^r
[2^{-(\betatild - \delta - \delta \epsilon)( \|k\normb - \|K\normb)} 2^{\betatild \|k\normb} \|p_k \|_{L^p}] \\
& ~~~~~~~~~~+
\sum_{K^*  < k } [2^{\betatild \|k\normb} \|p_k \|_{L^p}]
(2^{- (\betatild - \delta)\|k\normb} ) \\
& \lesssim
( 2^{-\betatild \|K\normb} 
+2^{- (\betatild - \delta) \|K^*\normb} )
\|f\|_{B^\beta_{p,q}} \\
&  \lesssim
N^{- \betatild} %D_{K,d}^{(1 - 1/q)_+}
\|f\|_{B^\beta_{p,q}}.
\end{align*}
This concludes the proof.
%=======================================
%Based on this decomposition, \cite{dung2011optimal} proposed an optimal adaptive recovery method 
%such that 
%the approximator has the form \eqref{eq:fNformat} under the conditions for $K,K^*,n_k$ given in the statement 
%and satisfies the approximation accuracy \eqref{eq:ffNoptimalAdaptiveApprox}.
%This can be proven by applying the proof of Theorem 3.1 in \cite{dung2011optimal} to the decomposition 
%\eqref{eq:DevorePopovExpansion} instead of Eq. (3.8) of that paper. See also Theorem 5.4 of \cite{dung2011optimal}.
%Moreover, the equivalence \eqref{eq:BpqEquivalence} gives the norm bound of the coefficient $(\alpha_{k,j})$.
\end{proof}

\begin{proof}[Proof of Proposition \ref{prop:BesovApproxByNN}]
We adopt the proof line employed by \cite{suzuki2018adaptivity}.
Basically, we combine Lemma \ref{lemm:Mnapproximation}
and 
Lemma \ref{lemm:BSplineInterpolation}.
We substitute the approximated cardinal B-spline basis $\check{M}$ into 
the decomposition of $f_N$ \eqref{eq:fNformat}.
%Let the approximated one be $\fcheck$.
Let the set of indexes $(k,j) \in \Integer \times \Integer$ 
that consists $f_N$ given in \Eqref{eq:fNformat}  
be $E_{N}$, i.e., $f_N = \sum_{(k,j)\in E_N} \alpha_{k,j} M_{k,j}^d$.
Accordingly, we set $\fcheck := \sum_{(k,j)\in E_N} \alpha_{k,j} \check{M}_{k,j}^d$.
Note that for each $x$, the number of $(k,j)\in E_N$ that satisfy ${M}_{k,j}(x) \neq 0$ is bounded by $(m+1)^d  (1+K^*) $,
and $\max_{(k,j)\in E_N} |\alpha_{k,j}| \lesssim 2^{ K^*  \frac{\betamin}{\betatild} (\betatild - 1/p)_+ }$ by the norm equivalence \Eqref{eq:BpqEquivalence}.
For each $x \in \Real^d$, it holds that
\begin{align*}
|f_N(x) - \fcheck(x)|
& \leq \sum_{(k,j) \in E_N} |\alpha_{k,j} |  |M_{k,j}^d(x) - \check{M}_{k,j}^d(x)| \\
& \leq \epsilon \sum_{(k,j) \in E_N} |\alpha_{k,j} |   \boldone\{M_{k,j}^d(x) \neq 0\} \\
& \lesssim \epsilon (m+1)^d 
(1+K^*) 2^{ K^*  (\betamin/\betatild) (\betatild - 1/p)_+ } \|f\|_{B^s_{p,q}} \\
& \lesssim \epsilon \log(N) 
N^{(1 + \nu^{-1} ) (\betatild - 1/p)_+}  \|f\|_{B^s_{p,q}}, %\\
%& \leq 
%\log(N) 
%N^{  2 d(1/p-1/r)_+/ (s - d(1/p-1/r)_+) + d^{-1}) (d/p-s)_+} \epsilon. % \\
%& \leq 
%\log(N) 
%N^{ - 2 d (1/p-1/r)_+ +  (d/p-s)_+/d} \epsilon
%N^{(d/p-s)_+}\|f\|_{B^s_{p,q}} \epsilon.
%\left(\sum_{k=0}^{K^*} 2^{s k q(1-d/p)}\right)^{1/q}
%K^* \epsilon \max |\alpha_{k,j} |
\end{align*}
where we used the definition of $K^*$ in the last inequality.
This evaluation yields that, for each $f \in U(B^\beta_{p,q}(\Omega))$, it holds that 
$$
\|f - \fcheck\|_{L^r}
\lesssim  
\|f - f_N\|_{L^r}
+
\|f_N- \fcheck\|_{L^r}
\lesssim \log(N) N^{(1 + \nu^{-1})(1/p-\betatild)_+}\|f\|_{B^s_{p,q}} \epsilon + N^{-\betatild}.
$$
By taking $\epsilon$ to satisfy 
$\log(N) N^{(1 + \nu^{-1})(1/p-\betatild)_+}\epsilon \leq N^{-\betatild}$, 
we obtain the approximation error bound.

As we have seen above $\max_{(k,j)\in E_N} |\alpha_{k,j}| \lesssim 2^{ K^*  \frac{\betamin}{\betatild} (\betatild - 1/p)_+ } \leq N^{(1 + \nu^{-1})(1/p-\betatild)_+}$.
The max of the absolute values of parameters used in $\check{M}_{k,j}^d$ can be bounded by $2^{K^*}$ (see \cite{suzuki2018adaptivity})
which is bounded by $N^{d (1 + \nu^{-1})(1/p-\betatild)_+}$.
Then, we obtain the assertion.

\end{proof}

\subsection{Proof of 
Theorem \ref{thm:AffineApproximationError} and 
Theorem \ref{thm:DeepApproximationError}}
\label{sec:ProofOfAffineDeepApprox}

\begin{proof}[Proof of Theorem \ref{thm:AffineApproximationError}]
This proof is almost obvious from Proposition \ref{prop:BesovApproxByNN}.
%The first layer $Ax + b$ can be realized by a neural network:
%Let $D = \|A x \|_\infty$, then 
%$$
%\eta(A x + D \boldone) = A x + D \boldone
%$$
%because all elements of $A x + D \boldone$ is non-negative.
%Therefore, we have that 
%$$
%Ax + b = \eta(A x + D \boldone) - (D \boldone + b).
%$$
We know that, from Proposition \ref{prop:BesovApproxByNN}, 
for $g \in U(B_{p,q}^\beta([0,1]^{\dtil}))$, there exists $\fcheck \in \Phi(L_1(\dtil),W_1(\dtil),S_1(\dtil),B_1(\dtil))$ such that 
$$
\|\fcheck - g\|_r \lesssim N^{-\betatild}.
$$
Because the density of the distribution of $Ax + b$ is bounded above when $x$ obeys the uniform distribution on $\Omega$, 
this also yields 
$$
\|\fcheck \circ (A \cdot + b) -g \circ (A \cdot + b)\|_r \lesssim N^{-\betatild}.
$$
(note that the Lebesgue measure on $\Omega = [0,1]^d$ corresponds to the uniform distribution on $\Omega$). %$L^r(\Omega)$ is the mean of $$).
If $\fcheck$ can be written as 
$$
\fcheck(x) = (\Well{L_1} \eta( \cdot) + \bell{L}) \circ \dots  
\circ (\Well{1} x + \bell{1}),
$$
then we have 
$$
\fcheck \circ (A \cdot + b)
= (\Well{L_1} \eta( \cdot) + \bell{L}) \circ \dots  
\circ (\Well{1} A \cdot  + \bell{1} + \Well{1} b) \in \Phi(L_1(\dtil),W_1(\dtil),S_1(\dtil),(\dtil C + 1) B_1(\dtil)).
$$
\end{proof}

\begin{proof}[Proof of Theorem \ref{thm:DeepApproximationError}]

\begin{align*}
\calHdeep := \{  h_H \circ \dots \circ h_1(x) \mid 
 h_\ell: [0,1]^{\mell} \!\! \to \! [0,1]^{\melle{\ell+1}},~h_{\ell,k} \in U(B_{p,q}^{\beta^{(\ell)}}([0,1]^{\mell}))~(\forall k \in [\melle{\ell+1}])\}.
\end{align*}

Since $\betatild^{(\ell)} > 1/p$, we can show that 
for each $h_{\ell,k}$, there exists $\fcheck_{\ell,k} \in \Phi(L_1(\mell),W_1(\mell),S_1(\mell),B_1(\mell))$ such that 
$$
\|\fcheck_{\ell,k} - h_{\ell,k} \|_\infty \lesssim N^{-\betatild}.
$$
Moreover, from the proof of Proposition \ref{lemm:BSplineInterpolation}, we can share all parameters other than the last layer among
$\fcheck_{\ell,k}~(k=1,\dots,\melle{\ell+1})$. If necessary, we may modify $\fcheck_{\ell,k}$ so that 
$0 \leq \fcheck_{\ell,k}(x) \leq 1~(\forall x \in [0,1]^{\mell})$ by adding one additional clipping layer which can be realized by ReLU
(actually, the clipping operator can be constructed by a linear combination of 2 nodes with ReLU activation as $f(x) = \max\{x,0\} - \max\{x - 1,0\} = \min\{\max\{x,0\},1\}$ for $x \in \Real$).
The approximation error of the whole layer can be evaluated as 
\begin{align*}
& \|h_{H} \circ \dots \circ  h_{1} - \fcheck_{H} \circ \dots \circ \fcheck_{1}\|_{\infty} \\
\leq 
&
\sum_{\ell=1}^H
\| h_{H} \circ \dots \circ h_{\ell+1} \circ h_{\ell} \circ \fcheck_{\ell-1} \circ \dots \circ  \fcheck_{1} -
h_{H} \circ \dots  \circ h_{\ell+1} \circ \fcheck_{\ell} \circ \fcheck_{\ell-1} \circ \dots \circ \fcheck_{1} \|_\infty \\
\leq &
\sum_{\ell=1}^H
\| h_{H} \circ \dots \circ h_{\ell + 1}  \circ h_{\ell}  
- h_{H} \circ \dots \circ h_{\ell + 1} \circ \fcheck_{\ell}
\|_\infty. %\circ \fcheck_{\ell-1} \circ  \fcheck_{1} -
%h_{H} \circ \dots \fcheck_{\ell} \circ \fcheck_{\ell-1} \circ  \fcheck_{1} \|_\infty \\
\end{align*}

Proposition \ref{lemm:EmbeddingLemma} tells that $h_{\ell',k} \in \calC^{(\betamin^{(\ell')} - 1/p) \wedge 1}$; thus,
$h_{\ell',k}$ is $\gamma_{\ell'}$-\Holder continuous where $\gamma_{\ell'} := (\betamin^{(\ell')} - 1/p) \wedge 1$.
Their composition $h_H \circ h_{H-1}\circ \dots \circ h_{\ell+1}$ is 
$\Gamma_\ell$-\Holder continuous where $B_\ell = \prod_{\ell' = \ell+1}^H \gamma_{\ell'}$.
Therefore, we have 
\begin{align*}
& \| h_{H} \circ \dots \circ h_{\ell + 1}  \circ h_{\ell}  
- h_{H} \circ \dots \circ h_{\ell + 1} \circ \fcheck_{\ell}
\|_\infty 
\lesssim
\| h_{\ell}  -  \fcheck_{\ell} \|^{B_\ell}_\infty,
\end{align*}
where $\|\cdot\|_\infty$ for a vector-valued function $g:\Real^{d'} \to \Real^{d''}$ is defined as
$\sup_{x} \|g(x)\|$.
Summing up this evaluation for $\ell=1,\dots,H$ concludes that 
$$
\|h_{H} \circ \dots \circ  h_{1} - \fcheck_{H} \circ \dots \circ \fcheck_{1}\|_{\infty} 
\lesssim \sum_{\ell=1}^H N^{- B_\ell \betatild^{(\ell)}}
\lesssim \max_{\ell \in H } N^{- \betatild^{*(\ell)}}.
$$
Consequently, the whole network can be realized as an element of 
$\Phi(L,W,S,B)$
where 
\begin{align*}
& L = \sum_{\ell=1}^H (L_1(\mell) + 1),~W = \max_\ell (W_1(\mell) \vee \melle{\ell+1}),\\
& S = \sum_{\ell=1}^H (S_1(\mell) + 3 \melle{\ell+1}),~
B = \max_\ell B_1(\mell).
\end{align*}
\end{proof}

\section{Proofs of estimation error bound (Theorem \ref{thm:EstimationErrorNNBesov_affine} and Theorem \ref{thm:EstimationErrorNNBesov_Deep}
)}
\label{sec:EstimationErrorAnalProof}

\begin{proof}[Proof of Theorem \ref{thm:EstimationErrorNNBesov_affine}]
We follow the proof strategy from \cite{2017arXiv170806633S,suzuki2018adaptivity} which uses Proposition \ref{prop:RiskBoundCovering}.
It suffices to 
the covering number of $\hat{\calF} = 
\{\fbar \mid f \in \Psi(L,W,S,B)\}$ for $(L,W,S,B)$ given in Theorem \ref{thm:AffineApproximationError}
where $\fbar$ is the clipped version of a function $f$.
Note that the covering number of $\hat{\calF}$ is not larger than that of $\Psi(L,W,S,B)$.
Hence, it is sufficient to evaluate that of $\Psi(L,W,S,B)$.
From Lemma \ref{lemm:CovNPhi}, the covering number of this class is upper bounded by
$$
\log N(\delta,\hat{\calF},\|\cdot\|_\infty)
\lesssim N \log(N) [\log(N)^2 + \log(\delta^{-1}) ].
$$

From Proposition \ref{prop:BesovApproxByNN}, there exists 
$\fcheck \in \Phi(L,W,S,B)$ such that
$$
\|\ftrue - R_K(\ftrue)\|_2 \lesssim N^{-\betatild}.
$$
Moreover, we notice that 
$
\|f - \ftrue\|_{\LPi(P_X)}^2 \leq R \|f - \ftrue\|_2^2.
$
for any $f:[0,1]^d \to \Real$ because the density $p_X$ of $P_X$ is bounded by $R$.
Therefore, by applying Proposition \ref{prop:RiskBoundCovering} with $\delta = 1/n$, we have 
\begin{align}
\EE_{D_n}[\|\fhat - \ftrue\|_{\LPi(P_X)}^2]
%\lesssim \EE_{D_n}[\|\fhat - \ftrue\|_2^2]
\lesssim N^{-2 \betatild} + \frac{N \log(N)(\log(N)^2 + \log(n)) }{n} + \frac{1}{n}.
\notag
%\label{eq:DnHatBound}
\end{align}
Here, we can minimize the right hand side by setting $N \asymp n^{\frac{1}{2\betatild + 1}}$ up to $\log(n)^3$-order, and then 
we obtain the estimation error of the least squares estimator as 
$$
n^{- \frac{2\betatild}{2\betatild + 1}} \log(n)^3.
$$
This yields the assertion.
\end{proof}

\begin{proof}[Proof of Theorem \ref{thm:EstimationErrorNNBesov_Deep}]
The proof is almost identical to the proof of Theorem 
\ref{thm:EstimationErrorNNBesov_affine}, except that we use Theorem \ref{thm:DeepApproximationError} as an approximation error bound.
\end{proof}
\section{Embedding theorem}
\label{sec:EmbeddingTheorem}

\begin{Lemma}
\label{lemm:Embedding}
For $0 < p^{(1)}, p^{(2)} \leq \infty$, let $\beta^{(1)}, \beta^{(2)} \in \Real_{++}^d$ such that 
they satisfy 
\begin{align}
& \tilde{\beta}^{(1)}  - \tilde{\beta}^{(2)} \geq \frac{1}{p^{(1)}} - \frac{1}{p^{(2)}}, \label{eq:BetaRelCond}\\ %~~~%\text{and}~~~
& \beta^{(2)} = \gamma \beta^{(1)},\notag \\
& p^{(1)} < p^{(2)}, \notag
\end{align}
for $0 < \gamma < 1$.
Then, it holds that 
$$
B_{p^{(1)},q}^{\beta^{(1)}} \hookrightarrow B_{p^{(2)},q}^{\beta^{(2)}}.
$$
\end{Lemma}
\begin{proof}
We show the assertion only for the situation where $p^{(1)} \neq \infty$, $p^{(2)} \neq \infty$, and $q \neq \infty$.
The proof for the setting in which $p^{(1)} = \infty$, $p^{(2)} = \infty$, or $q = \infty$ is satisfied is almost identical.
Recall the following norm equivalence shown in Lemma \ref{lemm:BSplineInterpolation}: 
$$
\|f\|_{B_{p,q}^s} \simeq 
\| (\alpha_{k,j})_{k,j}\|_{b^{\beta}_{p,q}} =  \left\{\sum_{k =0}^\infty \left[ 2^{k[\betamin - (\sum_{i=1}^d \lfloor k\beta'_i\rfloor/k) /p]}\Big(\sum_{j \in J(k)} |\alpha_{k,j}|^p\Big)^{1/p}  \right]^q\right\}^{1/q},
$$
when $p,q < \infty$.
Since $\frac{p^{(1)}}{p^{(2)}} < 1$, it holds that 
\begin{align*}
\Big(\sum_{j \in J(k)} |\alpha_{k,j}|^{p^{(1)}}\Big)^{1/p^{(1)}} 
& =
\Big(\sum_{j \in J(k)} |\alpha_{k,j}|^{p^{(2)} \frac{p^{(1)}}{p^{(2)}}} \Big)^{1/p^{(1)}} 
  \\
%2^{- (\sum_{i=1}^d \lfloor k\beta'_i\rfloor) 
%|J(k)|(\frac{1}{p^{(1)}} - \frac{1}{p^{(2)}}) } 
& \geq \Big(\sum_{j \in J(k)} |\alpha_{k,j}|^{p^{(2)}}  \Big)^{\frac{p^{(1)}}{p^{(2)}} \frac{1}{p^{(1)}} } 
=
\Big(\sum_{j \in J(k)} |\alpha_{k,j}|^{p^{(2)}}  \Big)^{\frac{1}{p^{(2)}} }.
\end{align*}
Moreover, we have 
\begin{align*}
& 2^{k[\betamin^{(1)} - (\sum_{i=1}^d \lfloor k\beta'^{{(1)}}_i\rfloor/k) /p^{(1)}]} \\
& \simeq 
2^{k \betamin^{(1)} - \sum_{i=1}^d \beta'^{{(1)}}_i /p^{(1)}}
= 
2^{k  \frac{\betamin^{(1)}}{\betatild^{(1)}} \left( \betatild^{(1)} - \frac{1}{p^{(1)}} \right) } 
\overset{(a)}{= }
2^{k  \frac{\betamin^{(2)}}{\betatild^{(2)}} \left( \betatild^{(1)} - \frac{1}{p^{(1)}} + \frac{1}{p^{(2)}} - \frac{1}{p^{(2)}}  \right) } \\
& 
\overset{(b)}{\geq}
2^{k  \frac{\betamin^{(2)}}{\betatild^{(2)}} \left( \betatild^{(2)} - \frac{1}{p^{(2)}}  \right) }
=
2^{k \betamin^{(2)} - \sum_{i=1}^d \beta'^{{(2)}}_i /p^{(2)}}
\simeq
2^{k[\betamin^{(2)} - (\sum_{i=1}^d \lfloor k\beta'^{{(2)}}_i\rfloor/k) /p^{(2)}]},
\end{align*}
where we used the condition $\beta^{(2)} = \gamma \beta^{(1)}$ in (a), and we used the condition from \Eqref{eq:BetaRelCond} in (b).
These relations yield the following evaluation:
\begin{align*}
&  \|f\|_{B_{p^{(1)},q}^{\beta^{(1)}}}  \\
& \simeq \| (\alpha_{k,j})_{k,j}\|_{b^{\beta^{(1)}}_{p^{(1)},q}} \\
& =  \left\{\sum_{k =0}^\infty \left[ 2^{k[\betamin^{(1)} - (\sum_{i=1}^d \lfloor k\beta'^{(1)}_i\rfloor/k) /p^{(1)}]}\Big(\sum_{j \in J(k)} |\alpha_{k,j}|^{p^{(1)}}\Big)^{1/p^{(1)}}  \right]^q\right\}^{1/q} \\
& \gtrsim \left\{\sum_{k =0}^\infty \left[ 2^{k[\betamin^{(2)} - (\sum_{i=1}^d \lfloor k\beta'^{(2)}_i\rfloor/k) /p^{(2)}]}\Big(\sum_{j \in J(k)} |\alpha_{k,j}|^{p^{(2)}}\Big)^{1/p^{(2)}}  \right]^q\right\}^{1/q} \\
& \simeq
\|f\|_{B_{p^{(2)},q}^{\beta^{(2)}}}.
\end{align*}
This yields the assertion.
\end{proof}

By combining Lemma \ref{lemm:Embedding} with 
the relation $B_{\infty,\infty}^{\gamma \beta} \hookrightarrow \calC^{\gamma\betamin}$ \citep{triebel2011entropy}, we immediately obtain the following corollary.

\begin{Corollary} %Lemma}
Suppose that $\betatild > p$, then for $\gamma = \frac{\betatild - p}{\betatild},$ it holds that
$$
B_{p,q}^{\beta} \hookrightarrow B_{\infty,q}^{\gamma \beta}
\hookrightarrow
B_{\infty,\infty}^{\gamma \beta}  \hookrightarrow \calC^{\gamma\betamin}.
$$
\end{Corollary}

\section{Minimax optimality}
\label{sec:MinimaxMixedSmooth}

In this section, we demonstrate the proof of Theorem \ref{thm:MinimaxRate}.
Before this, we prepare the basic notions.
The $\epsilon$-covering number $\calN(\epsilon,\mathcal{C},\hat{d})$ of a metric space $\mathcal{C}$ equipped with a metric $\hat{d}$
that is the minimal number of balls with radius $\epsilon$ measured by the metric $\hat{d}$ required to cover the set $\calC$
\citep{Book:VanDerVaart:WeakConvergence}.
Similarly, the $\delta$-packing number $\calM(\delta,\mathcal{C},\hat{d})$ is defined as the largest number of elements $\{f_1, \dots, f_{\calM} \} \subseteq \calC$ such that $\hat{d}(f_i,f_j) \geq \delta$ for all $i\neq j$.

\citet{raskutti2012minimax} showed the following inequality in their proof of Theorem 2(b) by utilizing the result by \citet{AS:Yang+Barron:99}.
\begin{Lemma}
\label{lemm:RaskuttiYangBarron:Minimax}
Let $\calF^\circ$ be the model of the true function.
For a given $\delta_n > 0$ and $\varepsilon_n > 0$, 
let $Q$ be the $\delta_n$-packing number $\calM(\delta_n,\calF^{\circ},\LPi(P_X))$ of $\calF^{\circ}$ and 
$N$ be the $\varepsilon_n$ covering number of that. %$\calN(\varepsilon_n,\calH_{\ell_{\infty}}^{d,q}(R),\LPi)$ of $\calH_{\ell_{\infty}}^{d,q}(R)$.
Suppose that they satisfy the following condition:
\begin{align}
& \frac{n}{2\sigma^2}\varepsilon_n^2 \leq \log(N),~\notag \\ %\label{eq:epsilonnbound}  \\
& 8 \log(N) \leq \log(Q),~  % \label{eq:NQbound} \\
4 \log(2) \leq \log(Q).   %\label{eq:2Qbound} 
\label{eq:QNboundMinimax}
\end{align}
Then, the minimax learning rate is lower bounded as 
\begin{align*}
\inf_{\fhat} \sup_{\fstar \in \calF^\circ}\EE_{D_n}[\|\fhat - \fstar \|_{\LPiPx}^2] 
\geq \frac{\delta_n^2}{4}.
\end{align*}
This concludes the assertion.
\end{Lemma}

Now, we are ready to show Theorem \ref{thm:MinimaxRate}.
\begin{proof}[Proof of Theorem \ref{thm:MinimaxRate}]
Proposition 10 of \cite{triebel2011entropy} showed that the $\epsilon$-covering number of the unit ball of anisotropic Besov spaces $B^{\beta}_{p,q}(\Omega)$ can be evaluated as 
\begin{equation*}
\log \calN(\epsilon,U(B^{\beta}_{p,q}(\Omega)), \|\cdot\|_r) \simeq \epsilon^{-1/\betatild},
%\label{eq:BesovCoveringNumCount}
\end{equation*}
for $0 < p,q \leq \infty$, $1 \leq r < \infty$, and $\beta \in \Real_{++}^d$ that satisfy
$$
\betatild > \max\left\{\frac{1}{p} - \frac{1}{r},\frac{1}{p} -1 ,0\right\}.
$$

{\bf Affine composition model}:

Apparently, $U(B^{\beta}_{p,q}(\Omega))$ is included in $\calHaff$. 
Hence, noting that $\Px$ is the uniform distribution and $\|\cdot\|_2 = \|\cdot\|_{\LPiPx}$, 
the covering number of $\calHaff$ can be lower bounded by 
$$
\log \calN(\calHaff, \|\cdot\|_{\LPiPx}) \gtrsim \epsilon^{-1/\betatild}.
$$
From this evaluation,
Lemma \ref{lemm:RaskuttiYangBarron:Minimax} yields that there exists $C_1 > 0$ independent of $n$ such that 
\begin{align*}
& \inf_{\fhat} \sup_{\fstar \in \calHaff} \EE_{D_n}[\|\fhat - \fstar \|_{\LPiPx}^2]  
\geq C_1 n^{-\frac{2\betatild}{2 \betatild + 1}}.
\end{align*}
To see this, we may just set $\epsilon_n \simeq \delta_n \simeq n^{-\frac{2\betatild}{2\betatild + 1}}$ in \Eqref{eq:QNboundMinimax} of Lemma \ref{lemm:RaskuttiYangBarron:Minimax}.

{\bf Deep composition model}:

Next, we show the minimax rate for the deep composition model.
Basically, we follow the same strategy developed by \citet{2017arXiv170806633S}, but we need to modify some technical details because we are dealing with anisotropic Besov spaces while \citet{2017arXiv170806633S} analyzed isotropic \Holder space.
Let $\ell^* := \min_{\ell \in [H]} \betatild^{*(\ell)}$, and $s^{(\ell)} := (\betamin^{(\ell)} - 1/p + \epsilon) \wedge 1$
where $\epsilon > 0$ can be arbitrary small for $q < \infty$ and $\epsilon = 0$ for $q = \infty$.
Without loss of generality, we may assume that $\beta_1^{(\ell)} \leq \beta_2^{(\ell)} \leq \dots \leq \beta_d^{(\ell)}$ for $\ell \in [H]$.
Let us consider a sub-model $\calHdeep'$ of $\calHdeep$ defined as   
\begin{align*}
\calHdeep' := & \{g_H \circ \dots \circ g_1  \mid \\
& g_\ell(x) = x~~(\ell = 1,\dots,\ell^*-1), \\
& g_{\ell^*}(x) = (g_{\ell^*,1}(x),0,\dots,0)^\top~~\text{where}~~g_{\ell^*,1} \in U(B^{\beta}_{p,q}(\Omega)), \\
& g_{\ell}(x) = (x_1^{ s^{(\ell)}},0,\dots,0)^\top~~(\ell = \ell^*+1,\dots,H) \}.
\end{align*}
%$$
%g_\ell(x) = x~~(\ell = 1,\dots,\ell^*-1),
%$$
%$$
%g_{\ell^*}(x) = (g_{\ell^*,1}(x),0,\dots,0)^\top 
%$$
%where $g_{\ell^*,1} \in U(B^{\beta}_{p,q}(\Omega))$, 
%and 
%$$
%g_{\ell}(x) = (x_1^{ s^{(\ell)}},0,\dots,0)^\top.
%$$
%As in \Eqref{eq:BesovCoveringNumCount}, we have that 
%$$
%aaa
%$$

For $\ell = \ell^* + 1,\dots, H$, through a cumbersome calculation, we can 
verify that $x_1^{ s^{(\ell)}} \in B_{p,q}^{\betamin^{(\ell)}}([0,1])$ for $x \in [0,1]$,
which ensures $g_{\ell,j}(x) \in B_{p,q}^{\beta^{(\ell)}}([0,1]^d)$ for $j = 1,\dots,d$.
To lower bound the covering number, we concretely construct a subset the cardinality of which can be easily estimated.
For that purpose, we use the expansion 
$f = \sum_{k=0}^\infty \sum_{j \in J(k)}  \alpha_{k,j} M_{k,j}^d(x)$ and the norm equivalence 
$\|f\|_{B_{p,q}^\beta} \simeq \| (\alpha_{k,j})_{k,j}\|_{b^\beta_{p,q}}$ given in Lemma \ref{lemm:BSplineInterpolation}.
For a while, we let $\beta := \beta^{(\ell^*)}$ and $B:=\prod_{q=\ell^* + 1}^H s^{(\ell)}$.
We define $k \in \Natural$ so that $k$ satisfies $2^{k \frac{\betamin}{\betatild}} \simeq n^{\frac{1}{1+2B \betatild}}$. %which will be set afterward.
For this choice of $k$, 
take a subset $\hat{J}(k) \subset J(k)$ such that $|\hat{J}(k)| \simeq |J(k)|$ and for each $j,  j' \in \hat{J}(k)$ with $j \neq  j'$,
the supports of $M_{k,j}^d$ and $M_{k,j'}^d$ are disjoint.
Using this index set $\hat{J}(k)$, we consider a set of functions that is given by
$$
\hat{\calH}_{\ell^*} := \left\{ f = \sum_{j \in \hat{J}(k)}  \alpha_{k,j} M_{k,j}^d(x) \mid \alpha_{k,j} \in \{0,2^{-k\betamin}\}\right\}.
$$
We can check that $|\hat{\calH}_{\ell^*}| = |\hat{J}(k)|\simeq 2^{k \sum_{j=1}^d \beta'_j} = 2^{k \betamin/\betatild}$
and $\|f\|_{B_{p,q}^{\beta}} \lesssim 1$ for all $f \in \hat{\calH}_{\ell^*}$ from the norm equivalence \eqref{eq:BpqEquivalence}.
For any $g_w = \sum_{j \in \hat{J}(k)}  w_j 2^{-k\betamin} M_{k,j}^d(x)  \in \hat{\calH}_{\ell^*}$~($w \in \{0,1\}^{|\hat{J}_k|}$), we can see that
\begin{align*}
f_w(x) & = g_{H}\circ \dots \circ g_{\ell^* + 1} \circ g_w \circ g_{\ell^* -1} \circ \dots \circ g_1(x) \\
& = \sum_{j \in \hat{J}(k)} w 2^{-B k\betamin} M_{k,j}^{dB} (x).
\end{align*}
If $w \neq w'$, then we can see that 
\begin{align*}
\|f_w - f_{w'}\|_{\LPiPx}^2 & \gtrsim \mathrm{Ham}(w,w') 2^{-2BK\betamin} 2^{-k \betamin/\betatild} \\
& \gtrsim \mathrm{Ham}(w,w') 2^{-k\betamin(2B\betatild + 1)/\betatild},
\end{align*}
where $\mathrm{Ham}$ is the Hamming distance because $\|M_{k,j}^{d}\|_{\LPiPx}^2 \simeq 2^{-k \betamin/\betatild}$.

Then, by the Varshamov--Gilbert bound (see Lemma 2.9 of \citet{tsybakov2008introduction}, for example),
there exists a subset $W_k \subset \{0,1\}^{|\hat{J}(k)|}$ such that $|W_k| \geq 2^{|\hat{J}(k)|/8}$ and 
$\mathrm{Ham}(w,w') \geq |\hat{J}(k)|/8$ for all $w, w' \in W_k$  with $w\neq w'$. This yields
$$
\|f_w - f_{w'}\|_{\LPiPx}^2 \gtrsim 2^{k\betamin/\betatild}  2^{-k\betamin(2B\betatild + 1)/\betatild} 
=  2^{-2 B k\betamin }  \simeq n^{- \frac{2 B \betatild}{2 B \betatild + 1}},
$$
where the definition of $k$ is used.
This implies that there exists a subset $\calHdeep'' \subset \calHdeep' (\subset \calHdeep)$ such that 
$$
\log(\calN(\epsilon_n,\calHdeep'',\|\cdot\|_{\LPiPx})) \gtrsim n^{\frac{1}{1+2B\betatild}}
$$
for $\epsilon_n \gtrsim n^{- \frac{B \betatild}{2 B \betatild + 1}}$.
Then, by Lemma \ref{lemm:RaskuttiYangBarron:Minimax}, we obtain that the minimax optima rate on $\calHdeep$ is lower bounded as 
\begin{align*}
\inf_{\fhat} \sup_{\fstar \in \calF^\circ}\EE_{D_n}[\|\fhat - \fstar \|_{\LPiPx}^2] 
\gtrsim n^{- \frac{B \betatild}{2 B \betatild + 1}}.
\end{align*}

\end{proof}

\section{Minimax optimal rate of linear estimators}
\label{sec:MinimaxLinearProof}

Define the convex hull of a function class $\calF^\circ$ as
$$
\mathrm{conv}(\calF^\circ) := 
\left\{ f(x) = \sum_{j=1}^M \lambda_j f_j(x) \mid 
M = 1,2,\dots,~f_j \in \calF^{\circ},~\lambda_j \geq 0,~
\sum_{j=1}^M \lambda_j = 1 \right\}.
$$
Let $\convbar(\cdot)$ is the closure of $\convbar(\cdot)$ with respect to $L_2(P_X)$-norm.

\begin{Proposition}[\cite{HayakawaSuzuki2019deep}]
\label{prop:ConvexHullArgument}
The minimax optimal rate of linear estimators on a target function class $\calF^\circ$ is the same as that on the convex hull of $\calF^\circ$:
\begin{align*}
\inf_{\text{$\fhat$: linear}} \sup_{\ftrue \in \calF^\circ}\EE_{D_n}[\|\ftrue -  \fhat\|_{\LPi(P_X)}^2]
=
\inf_{\text{$\fhat$: linear}} \sup_{\ftrue \in \convbar(\calF^\circ)}\EE_{D_n}[\|\ftrue -  \fhat\|_{\LPi(P_X)}^2].
\end{align*}
\end{Proposition}
See \cite{HayakawaSuzuki2019deep} for the proof of this proposition.

\begin{proof}[Proof of Theorem \ref{thm:MinimaxLinear}]

We basically follow the strategy developed by \cite{zhang2002wavelet}. 
Let $\mu$ be the uniform measure on $\Omega$.
They essentially showed the following statement in their Theorem 1.
Suppose that the space $\Omega$ has even partition $\calA$
such that $|\calA| = 2^{K}$ for an integer $K \in \Natural$, each $A$ has equivalent measure $\mu(A) = 2^{-K}$ for all $A \in \calA$,  and 
$\calA$ is indeed a partition of $\Omega$, i.e., $\cup_{A \in \calA} = \Omega$, $A \cap A' = \emptyset$ for $A,A' \in \Omega$ and $A \neq A'$.
Then, if $K$ is chosen as $n^{-\gamma_1} \leq 2^{-K} \leq n^{-\gamma_2}$ for constants $\gamma_1,\gamma_2 > 0$ that are independent of $n$, then there exists an event $\calE$ such that, for a constant $C' > 0$,
\begin{align*}
& |\{x_i \mid x_i \in A~(i\in \{1,\dots,n\})\}| \leq C' n/2^K~~(\forall A \in \calA), \\
& P(\calE) \geq 1 + o(1).
\end{align*}
We call this property of $\calA$ ``Condition A.''

Here, we consider a set $\calF^\circ$ of functions on $\Omega$ for which
there exists $\Delta > 0$ that satisfies the following conditions:
\begin{enumerate}
\item There exists $F >0$ such that, for any $A \in \calA$, there exists $g \in \calF^\circ$ that satisfies $g(x) \geq \frac{1}{2} \Delta F$ for all $x \in A$,
\item There exists $K'$ and $C''>0$ such that $\frac{1}{n}\sum_{i=1}^n g(x_i)^2 \leq C'' \Delta^2 2^{- K'}$ for any $g \in \calF^\circ$ on the event $\calE$.
\end{enumerate}
We call this condition of the function class $\calF^\circ$ ``Condition B.''

Let the minimax optimal rate of linear estimators on the function class $\calF^\circ$ be 
$$
R^* = \inf_{\fhat:\text{linear}} \sup_{\ftrue \in \calF^\circ} \EE_{D^n}[\|\fhat - \ftrue \|_{\LPiPx}^2].
$$
Then, under Conditions A and B,
there exists a constant $F_1$ such that at least one of the following inequalities holds:
\begin{subequations}
\label{eq:SufficientLinearMinimax}
\begin{align}
& \frac{F^2}{4 F_1 C''}  \frac{2^{K'}}{n}\leq  R^*, \label{eq:SufficientLinearMinimax1} \\ 
&\frac{ F^3}{32} \Delta^2 2^{-K} \leq R^*, \label{eq:SufficientLinearMinimax2} %\label{eq:SufficientLinearMinimax}
\end{align}
for sufficiently large $n$. 
\end{subequations}
%\phantom{\eqref{eq:SufficientLinearMinimax1},\eqref{eq:SufficientLinearMinimax2}}

{\it (i) Proof of \Eqref{eq:MinimaxLinearNonAdaptive}.}

For given $k \in \Natural$ (which will be fixed later), let $\Delta =  2^{- k[\betamin - (\sum_{i=1}^d \lfloor k\beta'_i\rfloor/k) /p]}$.
Then, from the wavelet expansion of anisotropic Besov space \eqref{eq:DevorePopovExpansion}, %it holds that 
$$
f_w =  \sum_{j \in J(k)} \Delta w_j  M_{k,j}^{d}(x) \in C U(B_{p,q}^\beta(\Omega)),
$$
where $C >0$ is a constant and $w = (w_j)_{j \in J(k)}$ is a one-hot vector, i.e., $w_j = 1$ for some $j \in J(k)$ and $w_{j'} = 0$ for all $j' \in J(k)$ with $j' \neq j$.
This expansion ensures that, for $K = {\sum_{i=1}^d \lfloor k\beta'_i\rfloor}$, there exists a partition $\calA$ of $\Omega$ that 
satisfies Condition A, and for any $A \in \calA$, there exists $w$ such that $f_w(x) \gtrsim \Delta$ for all $x \in A$ and 
$$
\frac{1}{n} \sum_{i=1}^n f_w(x_i)^2 \leq \frac{1}{n} \Delta^2 |\{i \mid x_i \in A~(i=1,\dots,n)\}| 
\lesssim \Delta^2 2^{-K},
$$
on the event $\calE$,
which ensures that $\calF^\circ = \{f_w \mid \text{$w$ is a one-hot vector}\}$ satisfies Condition B.
Hence, by choosing $k \in \Natural$ so that 
$2^{K} \simeq n^{\frac{1}{2 (\betatild + \frac{1}{2} - \frac{1}{p}) + 1}}$ (recall that $K = {\sum_{i=1}^d \lfloor k\beta'_i\rfloor}$ by definition), 
and setting $K=K'$, then 
\Eqref{eq:SufficientLinearMinimax} gives 
%  $K' = K = \sum_{i=1}^d \lfloor k\beta'_i\rfloor$
$$
R^* \gtrsim n^{-\frac{2 \betatild- v}{2\betatild- v + 1}},
$$
for $v = 2(1/p - 1/2)$. 
This yields the assertion because $\calF^\circ \subset C U(B_{p,q}^\beta(\Omega))$ for a constant $C$.

{\it (ii) Proof of \Eqref{eq:MinimaxLinearAffineHull}.}

Let $\betast := \betamax = \beta_1 = \dots = \beta_{\dtil} = \betamin$.
For $m$ such that $\betast < \min\{m,m-1+1/p\} $, let 
$\phi_{\dtil}(x) = \prod_{j=1}^{\dtil}\calN_m(x_i - (m+1)/2)$ ($x \in \Real^{\dtil}$).

{\bf (ii-a)} {\it Setting of $\dtil \geq d/2$:}

Let $V_{\dtil,d} :=\{U \in \Real^{\dtil \times d} \mid U  U^\top = \Id_{\dtil} \}$ be the Stiefel manifold and let $\pi_{V_{\dtil,d}}$ be the invariant measure on the Stiefel manifold (i.e., the uniform distribution).
Then, let $\bar{\phi}_{\dtil} : \Real^d \to \Real$ be 
$$
\bar{\phi}_{\dtil}(x) = \int \phi_{\dtil}(U x) \dd \pi_{V_{\dtil,d}}(U)~~~(x \in \Real^d).
$$
We can see that $\bar{\phi}_{\dtil}$ is spherically symmetric and there exists $F,C > 0$ such that 
$$
\bar{\phi}_{\dtil}(x) \geq F~~(\forall x\in \Real^d~\text{s.t.}~\|x\| \leq 1),
$$
and 
$$
\bar{\phi}_{\dtil}(x) \leq 
\begin{cases}
 C \|x\|^{-\dtil}~~~& (\|x\| \geq 1), \\
 1~~~& (\|x\| \leq 1).
\end{cases}
$$
The last inequality can be checked by the fact that for a sufficiently large $R > 0$, 
the measure of the set $\mu_R(\{ x \mid \|x\| = R,~\phi_{\dtil}(x) > 0 \}) \lesssim 1 \times R^{d - \dtil-1}/R^{d-1} = R^{-\dtil} $ (here, $\mu_R$ is the uniform probability measure on the sphere $S_{d-1}(R) = \{x \in \Real^d \mid \|x\| = R\}$) and $\|\phi_{\dtil}\|_\infty \leq 1$.

By the construction of $\phi_{\dtil}$ and the wavelet expansion of anisotropic Besov space \eqref{eq:DevorePopovExpansion} with the norm equivalence \eqref{eq:BpqEquivalence}, we have that there exists a constant $c > 0$ such that, for %any $b \in [0,1]^{\dtil}$ and 
any $k \in \Natural$ and $\bar{b} = \left[ \frac{1}{2} - 2^{-k}\left( \frac{m+1}{2} - \left\lfloor \frac{m+1}{2} \right\rfloor \right)\right]
(1,\dots,1)^\top \in \Real^{\dtil}$,
it holds that
$$
c \Delta \phi_{\dtil}\left(2^{k} (\cdot - \bar{b} )\right) %+ \frac{m+1}{2} \boldone_{\dtil}\right) 
\in U(B_{p,q}^{\betast} ([0,1]^{\dtil})),
$$
where $\Delta = 2^{-k (\betast - \dtil/p)}$. % and $\boldone_{\dtil} = (1,\dots,1)^\top \in \Real^{\dtil}$.
Here, let $0 < \bar{c} < 1$ be a constant such that $\bar{c}U(x-b') + \bar{b} \in [0,1]^{\dtil}$ for any $x,b' \in [0,1]^d$
and any $U \in V_{\dtil,d}$.  
Then, we have that, for any $b' \in [0,1]^d$, 
$$
c \Delta \phi_{\dtil}(2^{k} \bar{c} U ( \cdot -b') ) =  c \Delta \phi_{\dtil}(2^{k} ( \cdot - \bar{b})) \circ (\bar{c}U ( \cdot -b') + \bar{b}) \in \calHaff,
$$
for any $U \in V_{\dtil,d}$.
By the convex hull argument (Proposition \ref{prop:ConvexHullArgument}), this yields that 
$$
R_*^{\mathrm{lin}}(\calHaff) = R_*^{\mathrm{lin}}(\convbar(\calHaff)) \geq  
 R_*^{\mathrm{lin}}(\{c \Delta \bar{\phi}_{\dtil}(2^{k}  \bar{c} (\cdot -b')) \mid b' \in \Omega\}).
$$
Hence, it suffices to lower bound the far right-hand side of this inequality. 
We consider a partition $\calA$ of $\Omega$, where 
$A \in \calA$ has the form $A = [2^{-k} j_1, 2^{-k}(j_1 + 1)] \times \dots \times [2^{-k} j_d, 2^{-k}(j_d + 1)]$ for $0\leq j_i \leq 2^k -1$ ($i=1,\dots,d$).
Let $\hat{J}(k) = \{(j_1,\dots,j_d) \mid 0 \leq j_i \leq 2^{k-1}\}$ and $A_{j} = [2^{-k} j_1, 2^{-k}(j_1 + 1)] \times \dots \times [2^{-k} j_d, 2^{-k}(j_d + 1)] \in \calA$ for $j \in \hat{J}(k)$. Let $\varphi_{A_j} = c \bar{\phi}_{\dtil}(2^{k}  \bar{c} (\cdot -b_{A_j}))$, 
where $b_{A_j} = (2^{-k}(j_1 +1/2),\dots,2^{-k}(j_d +1/2))^\top$ for $j \in \hat{J}(k)$.
We can see that $|\calA| = 2^{d k}$. Hence, $\calA$ satisfies Condition A with $K = dk$ if $2^k$ is in polynomial order with respect to $n$.

Moreover, there exists $F > 0$ such that $\varphi_{A}(x) \geq F$ for all $x \in A$.
Next, we evaluate $\frac{1}{n} \sum_{i=1}^n \varphi_A(x_i)^2$.
On the event $\calE$, there exists $C'$ such that 
$|\{ i  \in [n] \mid x_i \in A' \}| \leq C' n/2^K = C' n \mu(A')$ for all $A' \in \calA$. 
Here, let 
$$
\bar{\varphi}_A(x) := 
\begin{cases} 
c C \|2^k\bar{c}(x - b_A)\|^{-\dtil} & (\|2^k\bar{c}(x - b_A)\| \geq 1), \\
 c &(\text{otherwise}),
\end{cases}
$$
then $\bar{\varphi}_A(x) \geq \varphi_A(x)$. Thus, we can upper bound $\frac{1}{n} \sum_{i=1}^n \varphi_A(x_i)^2$ as 
\begin{align*}
\frac{1}{n} \sum_{i=1}^n \varphi_A(x_i)^2
& \leq 
\frac{1}{n} \sum_{i=1}^n \bar{\varphi}_A(x_i)^2 
=
\frac{1}{n} \sum_{A' \in \calA } \sum_{ x_i \in A'} \bar{\varphi}_A(x_i)^2 
\leq 
\frac{1}{n} \sum_{A' \in \calA } C' \frac{n}{2^K} \max_{x \in A'} \bar{\varphi}_A(x)^2 \\
& = 
C' \sum_{A' \in \calA } \mu(A')  \max_{x \in A'} \bar{\varphi}_A(x)^2 
=
C' \sum_{A' \in \calA } \mu(A')  \min_{x \in A'} \bar{\varphi}_A(x)^2
\frac{\max_{x \in A'} \bar{\varphi}_A(x)^2 }{\min_{x \in A'}  \bar{\varphi}_A(x)^2} \\
&
\leq 
C' \sum_{A' \in \calA } \mu(A')  \min_{x \in A'} \bar{\varphi}_A(x)^2
\frac{\max_{x \in A'} \bar{\varphi}_A(x)^2 }{\min_{x \in A'}  \bar{\varphi}_A(x)^2} \\
%& C' \sum_{A' \in \calA } \mu(A')  \min_{x \in A'} \bar{\varphi}_A(x)^2
%\min\left\{
%\frac{ \max_{x \in A'} \|2^k\bar{c}(x - b_A)\|^{-2\dtil} }{\min_{x \in A'}   \|2^k\bar{c}(x - b_A)\|^{-2\dtil}},
%C^{-2} \max_{x \in A'}  \|2^k\bar{c}(x - b_A)\|^{2\dtil} \right\} \\ 
& \leq 
C' \sum_{A' \in \calA } \mu(A')  \min_{x \in A'} \bar{\varphi}_A(x)^2
\max_{x: \|2^k\bar{c}(x - b_A)\| \geq 1}
\frac{ \|2^k\bar{c}(x - b_A)\|^{-2\dtil} }{(\|2^k\bar{c}(x - b_A)\| + \bar{c} \|\boldone\|)^{-2\dtil}} \\
& \leq 
C' \sum_{A' \in \calA } \mu(A')  \min_{x \in A'} \bar{\varphi}_A(x)^2
(1 + \bar{c}\sqrt{d})^{2 \dtil} \\
& \leq C' (1 + \bar{c}\sqrt{d})^{2 \dtil} \int_{\Omega} \bar{\varphi}_A(x)^2 \dd x.
%\min\left\{
%\frac{ \max_{x \in A'} \|x - b_A\|^{-2\dtil} }{\min_{x \in A'}   \|x - b_A\|^{-2\dtil}},
%C^{-2} (1 + \bar{c}\sqrt{d})^{2\dtil} \right\} \\ 
\end{align*}
%
%this is upper bounded by $\int_{\Omega} \varphi_{A}(x)^2 \dd x $ (up to constant factor) which 
The quantity $\int_{\Omega} \bar{\varphi}_A(x)^2 \dd x$ on the right-hand side can be evaluated as 
\begin{align*}
& \int_{\Omega} \bar{\varphi}_{A}(x)^2 \dd x \leq \int_{x: \|x - b_A\| \leq 2 \sqrt{d}}  \bar{\varphi}_A(x)^2  \dd x \\
& \leq 
\int_{x: \|x - b_A\| \leq \bar{c}^{-1} 2^{-k}}  \bar{\varphi}_A(x)^2  \dd x + \int_{x:\bar{c}^{-1}   2^{-k} < \|x - b_A\| \leq 2 \sqrt{d}}  \bar{\varphi}_A(x)^2  \dd x \\
& \lesssim 
2^{- kd} + C \bar{c}^{-2 \dtil} 2^{-2 k \dtil} \int_{\bar{c}^{-1} 2^{-k} \leq r \leq 2 \sqrt{d}} r^{-2\tilde{d}} r^{d-1} \dd r \\
& \lesssim 
2^{- kd} + 2^{-2 k \dtil} \max\{2^{k(2\dtil - d)}, 1\} \\
& \lesssim
\max\{2^{-kd}, 2^{-2k\dtil}\}.
\end{align*}
Therefore, we have that, for a constant $C''$, on the event $\calE$, we have that 
$$\frac{1}{n} \sum_{i=1}^n \varphi_A(x_i)^2 \leq C'' (2^{-k d} \vee 2^{-2 k \dtil}).$$

Let $\calF^\circ = \{ \Delta \varphi_A \mid A \in \calA \}$, then $\calF^\circ$ satisfies Condition B.
%When $\dtil \leq d/2$, by choosing $k$ such that $\bar{c} 2^{k} \simeq n^{\frac{1}{2(\dtil + d/2 + \betast - \dtil/p)}}$, and setting $K=dk$ and $K' = 2 \dtil k$,  then through a simple calculation, \Eqref{eq:SufficientLinearMinimax} gives
%$$
%R_*^{\mathrm{lin}}(\calF^\circ) \gtrsim n^{-\frac{2(\betast - \dtil/p + d/2)}{2(\betast + \dtil - \dtil/p) + d}}.
%$$
When $\dtil \geq d/2$, by choosing $k$ so that $\bar{c} 2^k \simeq n^{\frac{1}{2(\betast + d - \dtil/p)}}$ and $K=K' = dk$, then \Eqref{eq:SufficientLinearMinimax} yields
$$
R_*^{\mathrm{lin}}(\calF^\circ) \gtrsim n^{-\frac{2(\betast - \dtil/p + d/2)}{2(\betast- \dtil/p + d/2) + d}}.
$$
This concludes the proof.

{\bf (ii-b)} {\it Setting of $\dtil < d/2$:}

Let $\calA$ be the partition of $\Omega$ as defined in the proof for $\dtil \geq d/2$, i.e., $|\calA| = 2^{dk}$ and each $A \in \calA$ can be written as $A = [2^{-k} j_1, 2^{-k}(j_1 + 1)] \times \dots \times [2^{-k} j_d, 2^{-k}(j_d + 1)]$ for $0\leq j_i \leq 2^k -1$ ($i=1,\dots,d$).
Pick up $A \in \calA$ and let $j \in \hat{J}(k)$ be the index such that $A = A_j$.
For a while, we fix $A$ and let $\bar{b} = b_{A_j} = (2^{-k}(j_1 +1/2),\dots,2^{-k}(j_d +1/2))^\top$ accordingly.
For $\theta = (w,b) \in \Real^{d - \dtil+1} \times \Real$, 
let $A_\theta: \Real^{d} \to \Real^{\dtil}$ be 
%$x \in \Real^d$, let 
$$
A_\theta(x) := 2^k [x_1 - \bar{b}_1,\dots,x_{\dtil-1} - \bar{b}_{\dtil-1}, w^\top (x_{\dtil:d} - \bar{b}_{\dtil:d})+b ],
$$
and consider 
\begin{align*}
\phi_\theta(x) := \phi_{\dtil}(A_\theta(x)).
%\phi_{\theta}(x) = 
%\prod_{j=1}^{\dtil-1} \calN_m(x_i - (m+1)/2) 
%\calN_m( \tau( x_i - (m+1)/2)
%(x \in \Real^{d}). 
\end{align*}
We take its convex hull with respect to $\theta$.
We note that 
$$
\phi_\theta(x) = \left(\prod_{j=1}^{\dtil -1} \calN_m(2^k(x_j - \bar{b}_j) - (m+1)/2) \right)  \calN_m\left(2^k [w^\top (x_{\dtil:d} - \bar{b}_{\dtil:d})+b]  - (m+1)/2\right). 
$$
To analyze its convex hull, it suffices to consider the convex hull of the last term $\calN_m\left(2^k [w^\top (x_{\dtil:d} - \bar{b}_{\dtil:d})+b]  - (m+1)/2\right)$.
Hence, we set $\psi(\cdot) := \calN_m(\cdot - (m+1)/2)$ and consider a set of functions %take its convex hull.
%Let 
$$
\tilde{\calF}_{C,\tau}^{(\psi)} := \{ x \in \Real^{d - \dtil + 1} \mapsto a \psi(\tau(w^\top  x + b))) \mid |a| \leq 2 C,~\|w\|\leq 1,~|b| \leq 2~(a,b \in \Real,~w \in \Real^{d - \dtil + 1})\}$$
for $C > 0$,~$\tau >0$.
We also define the Fourier transform of $\psi$ as $\hat{\psi}(\omega) := (2\pi)^{-1} \int e^{-\mathrm{i} \omega x} \psi(x) \dd x$ ($\omega \in \Real$).
Then, by Lemma \ref{lemm:GaussApprox}, we have that, for $h = 2^{-k}$ and $\tau = h^{-1-\kappa}$,  
\begin{align*}
& \inf_{\check{g} \in \convbar(\tilde{\calF}^{(\psi)}_{C,\tau})} \sup_{x \in [0,1]^d} 
\left|\check{g}(x) - \exp\left(- \frac{\|x - c\|^2}{2h^2}\right)\right|   \\ & 
\leq 
\frac{4}{|2\pi \hat{\psi}(1)|} 
\left[ C_{d-\dtil+1} R^{2(d-\dtil-1)}\exp(-R^2/2) + \exp(- R) \right],
\end{align*}
where $C = \frac{ \tau}{\pi |\hat{\psi}(1)|} = \Theta(h^{-1-\kappa})$ and $R = h^{-\kappa}(2\sqrt{d} + 1)$.
This indicates that, for a fixed $A \in \calA$, the convex hull of the set 
$
\{ a \phi_{\theta} \mid \theta = (w,b) \in \Real^{d - \dtil+1} \times \Real,~\|w\| \leq 1,~|b| \leq 2,~|a| \leq \Delta \}
$
where $\Delta = 2^{-k (\betast - \dtil/p)}$
contains $\varphi_A$ which satisfies 
\begin{align*}
& \left\|\varphi_A - 
\Delta (2C)^{-1} \left(\prod_{j=1}^{\dtil -1} \calN_m(2^k(x_j - \bar{b}_j) - (m+1)/2) \right) \exp\left( - \frac{\|x_{\dtil:d} - \bar{b}_{\dtil:d}\|^2}{2h^2}\right)
\right\|_\infty  \\
& = O\left(\Delta 2^{-k(1 + \kappa)} (h^{-\kappa(2(d - \dtil-1))} \exp(-h^{-2\kappa}/2) + \exp(-h^{-\kappa}))\right).
\end{align*}
We can see that on the event $\calE$, it holds that 
$$
\frac{1}{n} \sum_{i=1}^n \varphi_A^2(x_i) \lesssim \mu(A) (\Delta 2^{-k(1+\kappa)})^2 \lesssim 2^{-kd} 2^{-2k(\betast - \dtil/p + 1)} 2^{- 2k\kappa}= 2^{-2k(\betast - \dtil/p + 1 + d/2) -2k\kappa }, 
$$
for all $A\in \calA$. 
Let $\calF^\circ = \{ \varphi_A \mid A \in \calA \}$, then $\calF^\circ$ satisfies Condition B.
Note that, by the definition of $\tilde{\calF}_{C,\tau}^{(\psi)}$ is holds that $\varphi_A \in \convbar(\calHaff)$ for all $A \in \calA$. 
Thus
$$
R_*^{\mathrm{lin}}(\calHaff) = R_*^{\mathrm{lin}}(\convbar(\calHaff)) \geq  
 R_*^{\mathrm{lin}}(\calF^\circ).
$$

Therefore, by choosing $k$ such that $2^{k} \simeq n^{\frac{1}{2(\betast  - \dtil/p + 1 + d/2) + d + 2\kappa }}$, and setting $K=K'=dk$,  
then %by resetting $\kappa > 0$, 
\Eqref{eq:SufficientLinearMinimax} gives
$$
R_*^{\mathrm{lin}}(\calF^\circ) \gtrsim n^{-\frac{2(\betast - \dtil/p + 1 + \kappa + d/2)}{2(\betast - \dtil/p + 1 + \kappa + d/2) + d} }.
$$
\end{proof}

%For a sigmoid function $\sigma$, 
%Let %$\calF_{C,\tau}^{(\sigma)} := \{a \sigma(\tau(w  x + b)) \mid |a| \leq 2 C,~|w|\leq 1,~|b| \leq 1~(a,w,b \in \Real)\}$  
%and 
%$\tilde{\calF}_{C,\tau}^{(\psi)} := \{ x \in \Real^d \mapsto a \psi(\tau(w^\top  x + b))) \mid |a| \leq 2 C,~\|w\|\leq 1,~|b| \leq 2~(a,b \in \Real,~w \in \Real^d)\}$ for $C > 0$,~$\tau >0$.

\begin{Lemma}[\cite{suzuki2021benefit}]\label{lemm:GaussApprox}
%Let $\psi(x) = \frac{1}{2}(\sigma(x + 1) - \sigma(x - 1))$ and $\hat{\psi}$ be its Fourier transform: 
Let $h > 0$ and $R := h \tau/(2\sqrt{d} + 1)$.
Then, for $C = \frac{ \tau}{\pi |\hat{\psi}(1)|}$, 
the Gaussian RBF kernel can be approximated by 
\begin{align*}
& \inf_{\check{g} \in \convbar(\tilde{\calF}^{(\psi)}_{C,\tau})} \sup_{x \in [0,1]^d} 
\left|\check{g}(x) - \exp\left(- \frac{\|x - c\|^2}{2h^2}\right)\right|   \\ & 
\leq 
\frac{4}{|2\pi \hat{\psi}(1)|} 
\left[ C_d R^{2(d-2)}\exp(-R^2/2) + \exp(- R) \right]
\end{align*}
for any $c \in [0,1]^d$, where $C_d$ is a constant depending only on $d$.
In particular, the right hand side is $O(\exp(-n^\kappa))$ if $R = n^\kappa$.  
\end{Lemma}

\section{Auxiliary lemmas}

%Let the $\epsilon$-covering number 
%with respect to $\LPiPx$
%for a function class $\calG$ 
%be $\calN(\epsilon,\calG,\LPiPx)$
%as defined in the proof of Theorem \ref{eq:MinimaxOptimalboundOfmBesov}.
%Remind that it is defined as the minimal number of balls with radius $\epsilon$ measured by $\LPiPx$-norm needed to cover the set $\calG$ \citep{Book:VanDerVaart:WeakConvergence}.

The following proposition which were shown in \cite{2017arXiv170806633S,HayakawaSuzuki2019deep,AISTATS:Suzuki:2018} is convenient to show the estimation error rate.
\begin{Proposition}[\cite{2017arXiv170806633S,HayakawaSuzuki2019deep}]
\label{prop:RiskBoundCovering}
Let $\calF$ be a set of functions. Let $\fhat$ be the least-squares estimator in $\calF$: %any estimator in $\calF$. Define 
$$
%\Delta_n := \EE_{D_n}\left[\frac{1}{n}\sum_{i=1}^n (y_i - \fhat(x_i))^2 - \inf_{f\in \calF} \frac{1}{n}\sum_{i=1}^n (y_i - f(x_i))^2 \right].
\fhat = \argmin_{f \in \calF} \sum_{i=1}^n (y_i - f(x_i))^2.
$$
Assume that $\|\ftrue \|_\infty \leq F$ and all $f \in \calF$ satisfies $\|f\|_\infty \leq F$ for some $F \geq 1$.
If $\delta > 0$ satisfies $\calN(\delta,\calF,\|\cdot\|_\infty) \geq 3$, then
it holds that
$$
\EE_{D_n}[\|\fhat - \ftrue\|_{\LPi(P_X)}^2] 
\leq 
C \left[ \inf_{f \in \calF} \|f-\ftrue\|_{\LPi(P_X)}^2 + (F^2 + \sigma^2) \frac{\log \calN(\delta,\calF,\|\cdot\|_\infty) }{n} 
+ \delta (F + \sigma) \right],
%(1+\epsilon)^2 \left[ \inf_{f \in \calF} \|f-\ftrue\|_{\LPi(P_X)}^2 + F^2 \frac{56 \log \calN(\delta,\calF,\|\cdot\|_\infty) + 80}{n \epsilon} 
%+ 52 \delta F
%+ \Delta_n \right],
$$
where $C$ is a universal constant. 
%for any $\epsilon \in (0,1]$.

\end{Proposition}

The following lemma provides the covering number of the deep neural network model.
\begin{Lemma}[Covering number evaluation]
\label{lemm:CovNPhi}
The covering number of $\Phi(L,W,S,B)$ can be bounded by 
\begin{align*}
\log \calN(\delta,\Phi(L,W,S,B),\|\cdot\|_\infty) 
& \leq S \log(\delta^{-1} L (B \vee 1)^{L -1} (W+1)^{2L} ) \\
& \leq 2 S L \log( (B \vee 1)  (W+1) )
+
S  \log(\delta^{-1} L).
\end{align*}

\end{Lemma}

\begin{proof}[Proof of Lemma \ref{lemm:CovNPhi}]

Given a network $f \in \Phi(L,W,S,B)$ expressed as 
$$
f(x) =  (\Well{L} \eta( \cdot) + \bell{L}) \circ \dots  
\circ (\Well{1} x + \bell{1}),
$$
let 
$$
\calA_k(f)(x) = \eta \circ (\Well{k-1} \eta( \cdot) + \bell{k-1}) \circ \dots  \circ (\Well{1} x + \bell{1}),
$$
and 
$$
\calB_k(f)(x) =  (\Well{L} \eta( \cdot) + \bell{L}) \circ \dots  \circ (\Well{k} \eta(x) + \bell{k}),
$$
for $k=2,\dots,L$.
Corresponding to the last and first layers, we define $\calB_{L+1}(f)(x) = x$ and $\calA_{1}(f)(x) = x$ respectively.
Then, it is easy to see that $f(x) = \calB_{k+1}(f) \circ (\Well{k} \cdot + \bell{k}) \circ \calA_{k}(f)(x)$.
Now, suppose that a pair of different two networks $f, g \in\Phi(L,W,S,B)$ given by 
$$
f(x) = (\Well{L} \eta( \cdot) + \bell{L}) \circ \dots  
\circ (\Well{1} x + \bell{1}),~~
g(x) = ({\Well{L}}' \eta( \cdot) + {\bell{L}}') \circ \dots  
\circ ({\Well{1}}' x + {\bell{1}}'),
$$
has parameters with distance $\delta$: $\|\Well{\ell} - {\Well{\ell}}'\|_\infty \leq \delta$ and $\|\bell{\ell} - {\bell{\ell}}'\|_\infty \leq \delta$.
Now, not that $\|\calA_k(f) \|_\infty  \leq \max_j \|\Well{k-1}_{j,:}\|_1 \|\calA_{k-1}(f) \|_\infty + \|\bell{k-1}\|_\infty
\leq W B \|\calA_{k-1}(f) \|_\infty + B \leq (B \vee 1) (W +1 )\|\calA_{k-1}(f) \|_\infty 
\leq (B \vee 1)^{k-1} (W +1 )^{k-1}$,
and similarly, the Lipshitz continuity of $\calB_k(f)$ with respect to $\|\cdot\|_\infty$-norm is bounded as 
$
(B W)^{L - k + 1}.
$
Then, it holds that 
\begin{align*}
& |f(x) - g(x)| \\
= & \left|\sum_{k=1}^L \calB_{k+1}(g) \circ (\Well{k} \cdot + \bell{k}) \circ \calA_{k}(f)(x)
 - \calB_{k+1}(g) \circ ({\Well{k}}' \cdot + {\bell{k}}') \circ \calA_{k}(f)(x) \right| \\
\leq & \sum_{k=1}^L  (B W)^{L - k } \|
(\Well{k} \cdot + \bell{k}) \circ \calA_{k}(f)(x) - ({\Well{k}}' \cdot + {\bell{k}}') \circ \calA_{k}(f)(x)
 \|_\infty \\
\leq &  \sum_{k=1}^L  (B W)^{L - k }  \delta [W  (B \vee 1)^{k-1} (W +1 )^{k-1}  + 1] \\
\leq &  \sum_{k=1}^L  (B W)^{L - k }  \delta  (B \vee 1)^{k-1} (W +1 )^{k}
\leq \delta L (B \vee 1)^{L -1} (W+1)^{L}.
\end{align*}
Thus, for a fixed sparsity pattern (the locations of non-zero parameters), 
the covering number is bounded by $ \left( \delta/ [L (B \vee 1)^{L -1} (W+1)^{L}] \right)^{-S}$.
There are the number of configurations of the sparsity pattern is bounded by 
${(W+1)^L \choose S} \leq  (W+1)^{LS}$. Thus, the covering number of the whole space $\Phi$ is bounded as
$$
%\Phi(L,W,S,B) \leq 
 (W+1)^{LS} \left\{ \delta/ [L (B \vee 1)^{L -1} (W+1)^{L}] \right\}^{-S}
 = [\delta^{-1} L (B \vee 1)^{L -1} (W+1)^{2L} ]^S,
$$
which yields the assertion.

\end{proof}

%\newpage
%\appendix
%
%
%~\\
%
%\begin{center}
%{\bf \LARGE ------Appendix------}
%\end{center}
%
%\input{supplementary.tex}
%\input{./tex/supplementary.tex}

\end{document}